\documentclass[jair,twoside,11pt,theapa,letterpage]{article}
\usepackage{jair, theapa, rawfonts}

\usepackage{amsmath,amsfonts}
\usepackage{color,array}
\usepackage{cleveref}
\usepackage{graphicx}

\usepackage{caption}
\usepackage{subcaption}

\usepackage{multirow}
\usepackage{arydshln}

\usepackage{tabularx}
\usepackage{booktabs}

\usepackage{xcolor}

\usepackage{float}

\usepackage[xindy]{glossaries} 

\newacronym{nlp}{NLP}{Natural Language Processing}
\newacronym{llm}{LLM}{Large Language Model}
\newacronym{amr}{AMR}{Abstract Meaning Representation graphs}
\makeglossaries

\jairheading{82}{2025}{313-365}{06/2024}{1/2025}

\ShortHeadings{A Literature Review on the Challenges of Abstractive Dialogue Summarization}
{Kirstein, Wahle, Gipp \& Ruas}
\firstpageno{313}

\definecolor{myred}{RGB}{255, 54, 108}
\definecolor{myorange}{RGB}{254, 171, 59}
\definecolor{myblue}{RGB}{11, 223, 253}
\definecolor{mygreen}{RGB}{8, 197, 158}

\definecolor{red_2}{RGB}{245, 194, 213}
\definecolor{orange_2}{RGB}{251, 231, 163}
\definecolor{green_2}{RGB}{168, 248, 233}

\begin{document}

\title{CADS: A Systematic Literature Review on the Challenges of Abstractive Dialogue Summarization}

\author{\name Frederic Kirstein \email kirstein@gipplab.org \\
       \name Jan Philip Wahle \email wahle@uni-goettingen.de \\
       \name Bela Gipp \email gipp@uni-goettingen.de \\
       \name Terry Ruas \email ruas@uni-goettingen.de \\
       \addr Georg-August University Göttingen,
       Papendiek 14\\
       Göttingen, 37073, Germany}

\maketitle

\begin{abstract}
Abstractive dialogue summarization is the task of distilling conversations into informative and concise summaries.
Although focused reviews have been conducted on this topic, there is a lack of comprehensive work that details the core challenges of dialogue summarization, unifies the differing understanding of the task, and aligns proposed techniques, datasets, and evaluation metrics with the challenges.
This article summarizes the research on Transformer-based abstractive summarization for English dialogues by systematically reviewing 1262 unique research papers published between 2019 and 2024, relying on the \textsc{Semantic Scholar} and \textsc{DBLP} databases.
We cover the main challenges present in dialog summarization (i.e., language, structure, comprehension, speaker, salience, and factuality) and link them to corresponding techniques such as graph-based approaches, additional training tasks, and planning strategies, which typically overly rely on \textsc{BART}-based encoder-decoder models.
Recent advances in training methods have led to substantial improvements in language-related challenges. 
However, challenges such as comprehension, factuality, and salience remain difficult and present significant research opportunities.
We further investigate how these approaches are typically analyzed, covering the datasets for the subdomains of dialogue (e.g., meeting, customer service, and medical), the established automatic metrics (e.g., \textsc{ROUGE}), and common human evaluation approaches for assigning scores and evaluating annotator agreement.
We observe that only a few datasets (i.e., \textsc{SAMSum}, \textsc{AMI}, \textsc{DialogSum}) are widely used.
Despite its limitations, the \textsc{ROUGE} metric is the most commonly used, while human evaluation, considered the gold standard, is frequently reported without sufficient detail on the inter-annotator agreement and annotation guidelines.
Additionally, we discuss the possible implications of the recently explored large language models and conclude that our described challenge taxonomy remains relevant despite a potential shift in relevance and difficulty.
\end{abstract}

\section{Introduction}
\label{sec:introduction}
\begin{figure}[ht]
    \centering
    \includegraphics[width=0.99\linewidth]{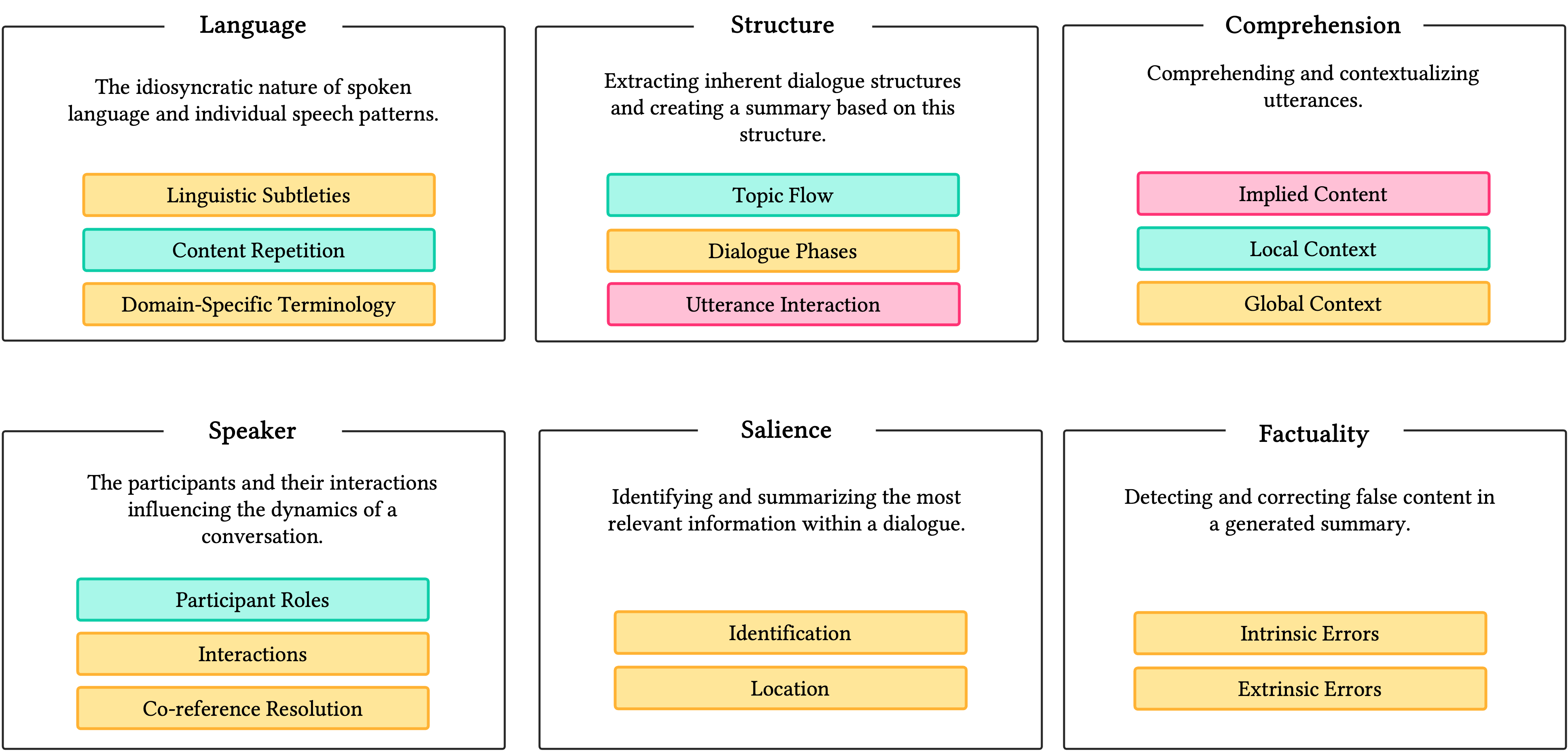}
    \caption{
    Overview of the six challenges in dialogue summarization, including a brief description of each challenge and an estimation of progress for related sub-challenges. Progress is evaluated based on two factors: (1) the extent to which mitigation strategies have been developed, and (2) the measurable improvement in summary quality as a result of these strategies. \colorbox{green_2}{Green} means mostly mitigated, \colorbox{orange_2}{orange} means good progress, and \colorbox{red_2}{red} stands for marked challenges still exist.
    }
    \label{fig:challenge_taxonomy}
\end{figure}

Abstractive dialogue summarization, a task within \gls{nlp} and text summarization, entails condensing key information from conversations into succinct and coherent summaries \shortcite{XuZZ22}.
This sub-field of text summarization is gaining prominence 
and is relevant for various real-world scenarios, including customer service (e.g., social media, \shortciteauthor{FeigenblatGSJ21}, 2021, and e-commerce, \shortciteauthor{LinZXZ22}, 2022), healthcare \shortcite{NairSK23}, daily life \shortcite{ChenLCZ21a}, meetings \shortcite{ManuvinakurikeSCN21}, and open-domain conversations (e.g., online-chat, \shortciteauthor{GliwaMBW19a}, 2019).
The relevance of this task, which offers insights into discussed concerns or issues, is thereby driven by the increase in digital conversations in these scenarios \shortcite{JonesRR04}.
While the common, manual approach to dialogue summarization is considered to be prone to human errors and consumes significant time and effort, automatic dialogue summarization may effectively reduce this overhead on the user side \shortcite{ManeKHP24}.

Diverging from traditional text summarization, which typically focuses on formal and linear content such as news articles \shortcite{RavautJC22} or scientific publications \shortcite{IbrahimAltmamiE22}, dialogue summarization demonstrates unique challenges \shortcite{KryscinskiKMX19}: conversational text is inherently dynamic, interactive, and non-linear \shortcite{SacksSJ74}, marked by verbosity, repetition, and informality.
Salient information is often distributed across multiple speakers, mixed with casual and off-topic remarks \shortcite{JiaRLZ22}.
Additionally, the use of elliptical and fragmented sentences, characterized by, e.g., incomplete thoughts and context-specific abbreviations, further hinders the summarization process.
These complexities make traditional text summarization approaches less effective, as they are not trained to bridge the gap between edited text and dialogue and struggle to transfer learned patterns through few-shot learning \shortcite{FengFQ22}.

Despite advancements achieved by leveraging and adapting general-purpose pre-trained neural language models \shortcite{LewisLGG20,RaffelSRL20}, and the recent trend to explore \glspl{llm} for this task \shortcite{LaskarFCB23}, there remains a notable gap in the field's foundational understanding of the challenges when processing conversation transcripts.
While research papers introduce techniques addressing similar challenges, their definitions and interpretations of these challenges can vary markedly.
For instance, while some describe the informal and ungrammatical nature of spoken language as a language challenge \shortcite{FengFQ22,RennardSHV23}, others frame the language challenge as different individual styles and unstructured expressions \shortcite{LeeYPS21,FangZCD22}.
Both viewpoints are thereby valid but, on their own, focus just on a subpart of the language challenge and blur the complete understanding of it.
Current surveys leave out the challenges in dialogue summarization and their variations in definitions, focusing instead on details on datasets, metrics, and techniques.
Consequently, there is no clear understanding of which aspects of dialogue summarization are well-addressed and where significant research potential lies.
Our systematic review consolidates the knowledge researched in the field considering Transformer-based models, aiming to provide a comprehensive overview of the challenges and their progress.
We further consider the interest in \glspl{llm} and discuss their impact on the relevance of the dialogue summarization challenges.

This article focuses on four key areas, namely challenges, dataset overviews, techniques, and evaluation methods.
In \textit{challenges}, we categorize known hurdles for Transformer-based models found when dealing with dialogue transcripts into six broader challenge blocks, i.e., language, structure, comprehension, speaker, salience, and factuality.
Our proposed taxonomy is displayed in Figure \ref{fig:challenge_taxonomy}, including the six challenge blocks, a short description, and exemplary sub-challenges of the challenge blocks.
In \textit{datasets overview}, our review displays used datasets focused on dialogue summarization, organized by their subdomain, and further an overview of techniques to generate datasets to cope with the existing data scarcity artificially and how to optimize data usage.
In \textit{techniques of dialogue summarization}, we then match 97 techniques proposed since 2019 to the six challenge blocks.
Furthermore, in \textit{evaluation methods}, we provide an overview of typical evaluation metrics used, spanning count-based metrics (e.g., \textsc{ROUGE}, \shortciteauthor{Lin04}, 2004), model-based (e.g., \textsc{BERTScore}, \shortciteauthor{ZhangKWW20}, 2020), QA-based (e.g., \textsc{QuestEval}, \shortciteauthor{ScialomDLP21}, 2021), and human evaluation metrics for performance and annotator agreement.
This systematic review contributes to a better understanding of the dialogue summarization problem, unifies the inherent definitions, overviews established techniques, and demonstrates the scarcity of datasets and fitting evaluation metrics.

\textbf{Organization of the Literature Review.}
The remainder of this review is structured as follows.
We introduce the current state of dialogue summarization in Section \ref{Sec:existing_overviews} and show our methodology in Section \ref{sec:methodology}.
In Section \ref{Sec:progress}, we provide the general problem definition and our challenges taxonomy consisting of six challenge blocks inherent in conversation scripts, further linking proposed approaches to handle the challenges.
Section \ref{Sec:datasets} overviews prominent datasets, while the display of evaluation approaches in Section \ref{Sec:evaluation} helps readers choose suitable indices to evaluate the effectiveness of a model.
Finally, Section \ref{Sec:discussion} discusses future research directions followed by conclusions in Section \ref{Sec:final_considerations}.
All resources for our review are publicly available.\footnote{https://github.com/FKIRSTE/LitRev-DialogueSum}

\subsection{Existing Reviews in Dialogue Summarization}
\label{Sec:existing_overviews}


Comprehensive literature reviews on dialogue summarization remain scarce, with notable contributions by \shortciteA{FengFQ22} and \shortciteA{JiaRLZ22}.
Existing surveys mainly explore the subdomains, their datasets, techniques, and metrics within dialogue summarization.
They provide a high-level overview and categorization of prominent subdomains like meeting, email, chat, medical, and customer service but often neglect niche areas such as e-commerce and debates \shortcite{FengFQ22}.
On the datasets, \shortciteA{TuggenerMDC21} provide a framework linking linguistic dialogue types \shortcite{WaltonK95} (e.g., persuasion, negotiation, inquiry) with established datasets, offering insights into dataset suitability.
\shortciteA{GuTL22} analyze dialogue summarization techniques up to early 2022, categorizing techniques by conversational context and component modeling (e.g., speaker, addressee, and utterance modeling to solve the `Who says what to whom' paradigm), covering a subset of the conversation challenges indirectly.
\shortciteA{JiaRLZ22} extend this analysis with techniques from 2023, exploring approaches like hierarchical models for long inputs \shortcite{ZhuXZH20a}, feature injection for enhanced understanding, auxiliary tasks for broader training objectives, and data augmentation from related tasks.
Regarding evaluations, works on dialogue summarization leverage automatic metrics from traditional document summarization, with the study by \shortciteA{GaoW22} discussing their individual strengths and limitations.

While existing surveys provide an understanding of dialogue summarization regarding subdomains, datasets, techniques, and metrics, the field misses a link between these categories through a common framework, such as the challenges inherent in processing dialogues.
Our review addresses this gap by introducing a comprehensive taxonomy of dialogue challenges and categorizing recent techniques until 2024 based on these challenges (Section \ref{Sec:progress}).
This approach highlights well-explored areas and points out gaps in current research. 
We analyze datasets (Section \ref{Sec:datasets}) and evaluation metrics (Section \ref{Sec:evaluation}), linking them to the identified challenges.
Additionally, we discuss the shift from smaller encoder-decoder models such as \textsc{BART} \shortcite{LewisLGG20} to \glspl{llm} and the implications for the relevance of the challenges with this new backbone model type (Section \ref{Sec:discussion}).
Our review focuses exclusively on dialogue summarization and does not encompass broader areas such as general text summarization \shortcite{RRMR23} or dialogue generation \shortcite{DengCSZ23}.

\section{Methodology}
\label{sec:methodology}

We use the \textsc{PRISMA} checklist \shortcite{PageMBB21} for organizing our literature review, as it is an established approach to set up a comprehensive systematic literature review while reducing the potential for incomplete data and biases in content selection and presentation \shortcite{Fagan17}.
This checklist ensures that our approach is structured, transparent, and reproducible.

\begin{figure*}
    \centering
    \includegraphics[width=0.95\linewidth]{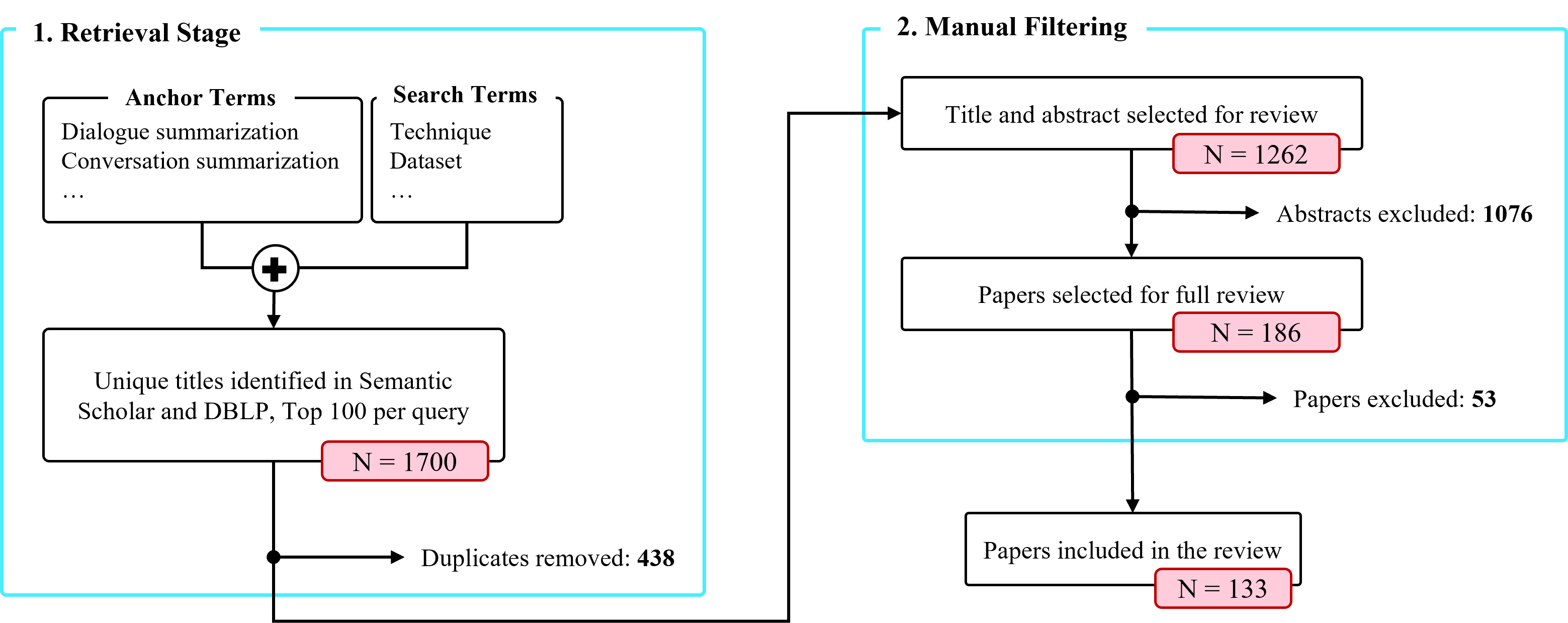}
    \caption{An overview of the different stages of our methodology and their inherent sub-stages. The \colorbox{red_2}{red} boxes indicate the number of papers considered in the respective step.}
    \label{fig:methodology_stages}
\end{figure*}

Our methodology comprises two stages, as detailed in Figure \ref{fig:methodology_stages}.

\noindent\textbf{Retrieval Stage.} We retrieve literature automatically and keyword-based from two established academic databases, i.e., \textsc{Semantic Scholar} and \textsc{DBLP}, using specific retrieval criteria and defining keyword queries for systematic searching.
The retrieved works are then automatically de-duplicated and saved to a list.

\noindent\textbf{Manual Stage.}
The second stage involves manually screening full documents using strict inclusion and exclusion criteria.
In this stage, we follow additional best practices for manual filtering beyond those in the \textsc{PRISMA} checklist. 
These include tracking specific items such as problem category, challenges discussed, and proposed technique category \shortcite{FoltynekMG19,SpindeHHR24} and guidelines for human annotators \shortcite{ParkS23,IbrahimS23}.

We evaluate 1262 unique papers from 2019 until 2024, offering an up-to-date perspective on the state of the art in dialogue summarization using Transformer-based architectures.
We decided against including references from selected papers that did not appear in our crawl as additional sources to maintain an unbiased selection process and avoid complicating the stopping criterion for literature inclusion.
Considering our exclusion criteria, such as multilingualism, non-abstractive, and non-Transformer-based approaches, we select 133 papers for our literature review.

\subsection{Stage 1: Retrieving Candidate Documents}
\label{sec:methodology_crawl}

\noindent \textbf{Information Sources.}
Our primary data sources are \textsc{Semantic Scholar}\footnote{https://www.semanticscholar.org/} and \textsc{DBLP}\footnote{https://dblp.org/} (DataBase systems and Logic Programming), both recognized for extensive coverage in computer science \shortcite{Kitchenham04,BreretonKBT07a}.
While \textsc{Semantic Scholar} allows for advanced search functionalities and complex queries \shortcite{XiongLCL18,Hannousse21,WangY21}, DBLP is the most comprehensive database for computer science publications, frequently used in other surveys as sole source \shortcite{NguyenJCD21,DongLGC22,ZhouXTZ22} and encompassing major peer-reviewed journals and conference proceedings.
This dual-source strategy ensures thorough retrieval and coverage specific to dialogue summarization, strengthening the certainty and reliability of our evidence base.
To ensure our literature review remains up-to-date, we regularly re-crawl databases while conducting the review to incorporate the latest publications into our analysis.\footnote{Last crawled on March 25, 2024.}
While we acknowledge the potential for these databases to miss the most recent uploads, our strategy minimizes this risk.
We exclude databases such as \textsc{Google Scholar}, which overly returned gray literature, and \textsc{Web of Science}, which emphasized extractive summarization techniques during our pre-testing, to maintain a focus on peer-reviewed literature in abstractive dialogue summarization.
We design an automatic pipeline to retrieve query-related works and remove duplicates.
We limit the number of results per query to the top 100 based on the relevance ranking provided by the platforms, which was rarely required.
We recognize that the ranking systems may be biased towards prestigious journals or highly cited authors.
Hence, we retrieve all papers published after June 2023, not relying on the ranking systems to account for their limited time to gather citations.

\noindent \textbf{Data Collection Process.}
\label{subsec:retrieving_candidates}
Our search strategy uses key terms related to dialogue summarization identified during our initial related work analysis (Section \ref{Sec:existing_overviews}).
Anchor terms such as `dialogue summarization' and `multi-party conversation' are combined with specific search terms such as `technique.'
We use four anchor terms and 28 search terms, resulting in 112 search queries for each database.
We present the complete list and search keywords in our repository\footnote{https://github.com/FKIRSTE/LitRev-DialogueSum \label{repository_link}}.
Employing a Python-based pipeline (Section \ref{repository_link}), we systematically extract documents from \textsc{Semantic Scholar} and \textsc{DBLP} APIs, merging and unifying the results into tabular data.
Between January 2019 and March 2024, we retrieved 732 publications from \textsc{DBLP} and 968 from \textsc{Semantic Scholar}.
After removing 438 duplicates identified across both sources, we conclude the stage with 1262 unique publications.
These results are tagged with their respective queries and compiled into a CSV file for subsequent selection and analysis.

\subsection{Stage 2: Manual Filtering}
\label{sec:methodology_screening}

\noindent \textbf{Eligibility Criteria.}
\label{subsec:criteria}
For our manual filtering process, we define eligibility criteria to ensure the focus of our literature review.
We consider open-access papers published since 2019 that align with the rise of Transformer-based text summarization models such as \textsc{BART} \shortcite{LewisLGG20} and \textsc{PEGASUS} \shortcite{ZhangZSL20}, which have been the core model for most approaches in the field.
This timeframe further allows us to concentrate on current challenges rather than including some that may have lost relevance (e.g., negation, \shortciteauthor{KhalifaBM21}, 2021) because of advancements in the field.
We exclude non-English primary datasets, multi-modal studies, and extractive or non-Transformer-based methods and focus on English-language research on abstractive dialogue summarization.

\noindent \textbf{Manual Paper Processing.}
A team of three reviewers\footnote{Each team member discloses potential conflicts of interest to ensure an unbiased review process.}, consisting of postgraduate students and doctoral candidates with a computer science background, undertake the manual screening process, starting with titles and abstracts to narrow down the automatic selection from 1262 to 186 documents using our eligibility criteria.
They then review the full texts, checking if the works still meet the eligibility criteria, and decide whether to include, exclude, or discuss each paper.
Each paper undergoes a secondary review to minimize bias, reaffirming initial selections and categorizations, thereby enhancing the study's integrity.
Eventual disagreements are resolved through consensus.
The annotators also track features such as used datasets, employed backbone models, proposed techniques, and evaluation approaches.
This step filters out 53 papers, leaving 133 to compose the literature review and detail challenges, techniques, datasets, and evaluation metrics.

\section{Challenges and Progress} 
\label{Sec:progress}
Abstractive dialogue summarization aims to condense multi-party conversations into their key points, ranging from casual chats to expert discussions, and whether these points were explicitly stated or inferred \shortcite{CohanDKB18}.
The task was first thoroughly defined in 2002 by \shortciteauthor{Zechner02}, outlining challenges unique to spoken dialogues such as handling speech disfluency, the lack of clear sentence boundaries, distribution of salient information across various speakers and turns, resolving references, identifying discourse structures, and dealing with inaccuracies caused by speech recognition errors.
While some definitions take a mathematical approach \shortcite{JiaRLZ22}, viewing dialogues as sequences of turns that are compressed and associated with a specific speaker, other definitions adopt a broader description of the task as `extracting useful information from a dialogue' \shortcite{LiuWXL19}.
Although all definitions contain relevant, overlapping aspects, they have different viewpoints, each considering a different set of challenges to define dialogue summarization.
Our goal is to organize these definitions and provide a concise, structured overview of the progress in abstractive dialogue summarization, extending the earlier challenge-based concept \shortcite{Zechner02} and proposing the \textsc{Challenges of Abstractive Dialogue Summarization (CADS)} Taxonomy (\Cref{fig:challenge_taxonomy}) tailored to the current Transformer-based model architecture.
This taxonomy comprises six challenges: language, structure, comprehension, speaker, salience, and factuality.
The first five challenges are related to the input, while the last concerns reliability.
To develop this comprehensive framework, we employed human annotators (\Cref{sec:methodology_screening}) to systematically analyze and categorize recurring challenges in the literature and consolidate them into the six challenges.
Each pillar encompasses related sub-challenges that represent key focus areas consistently addressed in the literature along the broader challenges.
In the following \Cref{sec:challenge_language,sec:challenge_structure,sec:challenge_understanding,sec:challenge_speaker,sec:challenge_salience,sec:challenge_factuality}, we detail the CADS Taxonomy as shown in \Cref{fig:challenge_taxonomy}.
An overview of the approach strategies proposed to mitigate the individual challenges is shown in \Cref{tab:overview_approaches}.
\Cref{tab:expected_error_types} summarizes the expected errors when challenges are not successfully mitigated, using the analysis by \shortciteA{KirsteinWRG24} as a base for the individual error types.

For this section, we consider 91 papers, from which 73 papers define the challenge characteristics (sub-challenges in \textit{italic}) and 74 describe explored techniques.
We exclude papers mainly discussing evaluation strategies and datasets from this subset and cover them in \Cref{Sec:datasets,Sec:evaluation}.

\newcolumntype{L}[1]{>{\raggedright\arraybackslash}p{#1}}
\begin{table}[t]
    \centering
    \small
    \begin{tabular}{L{2.5cm} L{4cm} L{7.5cm}}
         \toprule
         Challenge & Approach Category & Related Papers \\
         \midrule
         
         \multirow{1}{*}{Language} 
         & Pre-training & \shortciteA{RaffelSRL20,ZouZHG21,ZhouLCL23,LyuPLB24} \\
         & Training tasks & \shortciteA{ZhuXZH20a,KhalifaBM21,LeeYPS21,JiaLTZ22,BertschNG22} \\
         \cmidrule(lr){2-3}
         & Pre-processing & \shortciteA{GaneshD19} \\
         \midrule
         
         \multirow{1}{*}{Structure} 
         & Pre-training & \shortciteA{LeeKHU21,PeysakhovichL23,XuJK24} \\
         & Training tasks & \shortciteA{FengFQQ21,LeeYCJ21,LiuZZC21,YangWTW22} \\
         \cmidrule(lr){2-3}
         & Architecture modification & \shortciteA{Li22,LeiZYH21,GaoCLC23,HuaDM23,HuaDX22} \\
         & Importance measures & \shortciteA{ReimersG19a,LiangWCB23a} \\
         \midrule
        
        \multirow{1}{*}{Comprehension} 
         & Architecture modification & \shortciteA{zotero-18877} \\
         \midrule

        \multirow{1}{*}{Speaker}          
         & Training tasks & \shortciteA{GanZKY21,QiLFL21,AsiWEG22,NarakiSH22} \\
         \cmidrule(lr){2-3}
         & Architecture modification & \shortciteA{LeiYZH21,LeiZYH21,LiuSC21,HuaDX22} \\
         & Pre-processing & \shortciteA{JoshiCLW20a,LeeLWL21}\\
         & Post-processing & \shortciteA{FangZCD22,LiuC22} \\
         \midrule

        \multirow{1}{*}{Salience}          
         & Pre-training & \shortciteA{PagnoniFKW23,ZhangLYF23} \\
         & Training tasks & \shortciteA{ChauhanRDG22,LiuZXL22a,10.1016/j.eswa.2021.116292} \\
         \cmidrule(lr){2-3}
         & Architecture modification & \shortciteA{LiXYW21,HuaDM23} \\
         & Human feedback & \shortciteA{ChenDY23} \\
         & Loss function & \shortciteA{HuangSMX23} \\
         & Pre-processing & \shortciteA{LiuC21,JungSJC23} \\
         \midrule
        
        \multirow{1}{*}{Factuality} 
         & Training tasks & \shortciteA{GanZKY21,TangNWW22} \\
        \cmidrule(lr){2-3}
         & Architecture modification & \shortciteA{WuLLS21,ZhaoZXG21,ZhaoZHZ21,NairSK23} \\
         & Human feedback & \shortciteA{ChenDY23} \\
         & Loss function & \shortciteA{LiuZXL22a,HuangSMX23} \\
         & Post-processing & \shortciteA{FuZWL21,LiWGS23} \\
         \bottomrule
    \end{tabular}
    \caption{Overview of challenges, major approach trends, and corresponding literature. More details are stated in the article's accompanying repository: \\ https://github.com/FKIRSTE/LitRev-DialogueSum.}
    \label{tab:overview_approaches}
\end{table}

\subsection{Language}
\label{sec:challenge_language}

\begin{figure}
    \centering
    \includegraphics[width=0.9\linewidth]{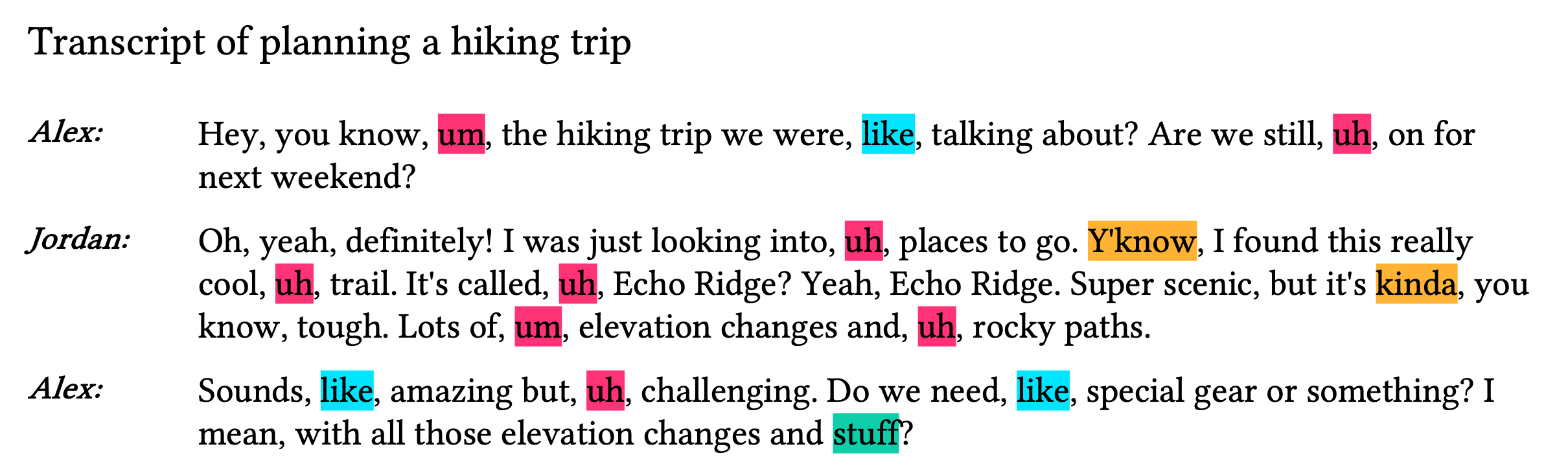}
    \caption{Dialogue snippet showing examples of the idiosyncratic nature of spoken language and individual speech patterns. \colorbox{myred}{Red} displays disfluencies, \colorbox{myblue}{blue} highlights personal speech patterns, \colorbox{myorange}{orange} stands for colloquialism, \colorbox{mygreen}{green} represent informal expressions.}
    \label{fig:challenge_language}
\end{figure}

\noindent \textbf{Characteristics.}
The language challenge describes the idiosyncratic nature of spoken language and individual speech patterns.
It includes \textit{linguistic subtleties} such as informal expressions (e.g., `yeah,'), ungrammatical structures, colloquialisms, personal vocabulary, and linguistic noise such as filler words \shortcite{KoayRDB20d,ZhangNYZ21a,FengFQ22,KumarK22,AntonyASK23}.
It further covers \textit{content repetition}, i.e., speakers restate or rephrase information for emphasis or clarity \shortcite{ChenY20,KhalifaBM21,LeiZYH21} and \textit{domain-specific terminology}, e.g., medical terms \shortcite{BertschNG22,LiuZXL22,LiuZXL22a,Li22}.
These elements demand that models accurately interpret and adapt to the linguistic features of dialogues \shortcite{JiaLTZ22}. 
\Cref{fig:challenge_language} presents a dialogue snippet showing examples of these three sub-challenges.
Failing to handle this challenge may result in a loss of coherence and clarity, repetition of content, and factuality issues.

\noindent \textbf{Approaches.}
Transformer-based models often struggle with the language characteristics due to a gap between their pre-training data, typically well-edited texts (e.g., news articles, research papers, Wikipedia entries, \shortciteauthor{RaffelSRL20}, 2020), and the characteristics of spoken dialogue not reflected in these edited texts \shortcite{ZouZHG21}.
The transfer between these text styles is hindered by the scarcity of diverse, dialogue-oriented pre-training data, resulting in a lack of exposure to dialogue data during training, consequently reducing models' performance.
Repetition of already provided information is less of an issue for current language models \shortcite{KhalifaBM21}.

Current research aims to bridge this gap between the formal language of the pre-training data and dialogues by retraining models with dialogue-focused tasks, considering both single- and multi-task setups.
Explored tasks span masking key dialogue elements (e.g., pronouns entities, high-content tokens, and words, \shortciteauthor{KhalifaBM21}, 2021) and part-of-speech tagging \shortcite{LeeYPS21}.
More recent tasks aim to transform informal, first-person dialogue into a structured, third-person narrative through changing speaker names, adjusting grammar, and adding emotional context \shortcite{BertschNG22}, or simulating a dialogue by modifying pre-training documents into a conversation structure \shortcite{ZhuXZH20a}.
Teaching models to understand the `who-did-what' structure through a specific task is further gaining traction \shortcite{JiaLTZ22}.
Other explored strategies include pre-processing the input for anaphora resolution \shortcite{GaneshD19} and adapted pre-training, e.g., to adjust to specific terminology \shortcite{ZouZHG21,ZhouLCL23,LyuPLB24}.

\subsection{Structure}
\label{sec:challenge_structure}

\begin{figure}
    \centering
    \includegraphics[width=1.0\linewidth]{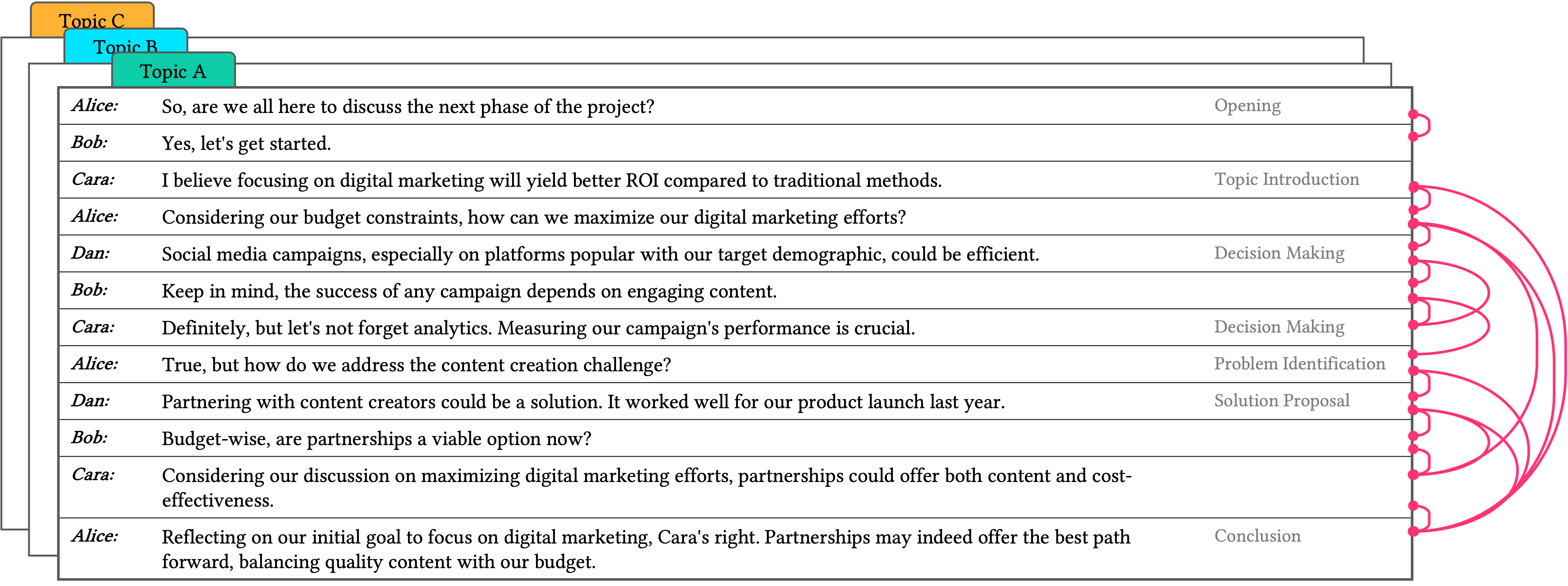}
    \caption{Excerpt from a conversation to demonstrate the structural challenge, with a detailed view on one topic out of multiple topics touched in the whole conversation. The red arcs mark the connection between individual turns, demonstrating how different dialogue phases (grey text), such as the conclusion, relate to many previous conversation phases.}
    \label{fig:challenge_structure}
\end{figure}

\noindent \textbf{Characteristics.}
Aiming to extract a conversation's built-up, the structure challenge is about segmenting a dialogue transcript and creating an ordered summary based on the extracted structures \shortcite{MaZGW23}.
This challenge involves identifying the \textit{topic flow} of a conversation \shortcite{ZhangZ21,ZhangNYZ21a,GuTL22,ShindeGSB22}, tracking the \textit{dialogue phases} (e.g., problem identification, decision making, \shortciteauthor{TuggenerMDC21}, 2021; \shortciteauthor{LiHXA23a}, 2023), and analyzing the \textit{utterance dependency} (i.e., relationship and dependencies between utterances, \shortciteauthor{FengFQ22}, 2021a; \shortciteauthor{LeiZYH21} 2021b; \shortciteauthor{HuaDX22}, 2022).
The dialogue in \Cref{fig:challenge_structure} shows topic flow, dialogue phases, and utterance dependency.
This challenge can lead to summaries lacking coherence, completeness, and depth \shortcite{LeeLWL21,LiuZZC21,QiLFL21,YangWTW22,LiangWCB23a}.

\noindent \textbf{Approaches.}
Transformer-based models face two primary hurdles when handling the structural challenge.
First, they must adapt to the variability in dialogue structures, such as differing topic sequences and conversation phases depending on the conversation type. 
Second, they need to track and connect long-range dependencies to link utterances.
The first difficulty is related to the lack of varied datasets across different dialogue types on which models can train to improve generalization as well as that most pre-trained models used are expecting more precise structural signals due to the more structured data used during pre-training (\Cref{sec:challenge_language}).
The second problem relates to a weakness of transformed-based architectures: their limitation in handling long sequences, which stems from the practical limitation of a fix-sized context window \shortcite{LeeKHU21}, the blurred attention on a single token when many tokens are processed \shortcite{XuJK24}, and the bias towards recent information \shortcite{PeysakhovichL23}.

Recent methods employ graph structures to understand and use dialogue structures more effectively.
They combine static and dynamic graphs for a detailed examination of conversation dynamics \shortcite{GaoCLC23} where the static graphs represent unchanging aspects like speaker relationships, and dynamic graphs track how dialogues evolve.
Abstract Meaning Representation (AMR) graphs are employed to capture overarching themes \shortcite{HuaDM23}, detailed sentence-level connections, and entity interactions, enhancing content comprehension \shortcite{HuaDX22}.
Meanwhile, a hybrid approach combining traditional language models with graph neural networks captures the utterance dependency and addresses the long-range dependency sub-challenge \shortcite{Li22}.
Similarly, ConceptNet \shortcite{LiuS04} is used for thematic linking \shortcite{LeiZYH21}.

To manage long dialogue transcripts beyond context-size limits and the inherent long-distance relations, researchers are moving away from simply truncating inputs to context length to avoid losing vital information.
An established approach segments the summarization process into stages (e.g., topics and dialogue phases) for coherent processing \shortcite{LaskarFCB23,AsthanaHHH24,MullickBRK24}.
The segmentation and multi-pass strategy iteratively summarizes parts of the dialogue and combines partial summaries with a subset of the remaining content in each iteration to build a complete summary \shortcite{SharmaFJ23}.
The $Summ^N$ framework \shortcite{ZhangNMW21} summarizes dialogue segments independently before refining them together, offering a scalable solution.

Other techniques involve additional training tasks \shortcite{FengFQQ21,LeeYCJ21,LiuZZC21,YangWTW22}, sentence importance measures to group sentences into sub-topics \shortcite{ReimersG19a,LiangWCB23a}, while hierarchical setups to differentiate dialogue parts based on who is speaking (word level) and what is being discussed (turn level) are also popular \shortcite{ZhuXZH20a}.

\begin{figure}
    \centering
    \includegraphics[width=0.9\linewidth]{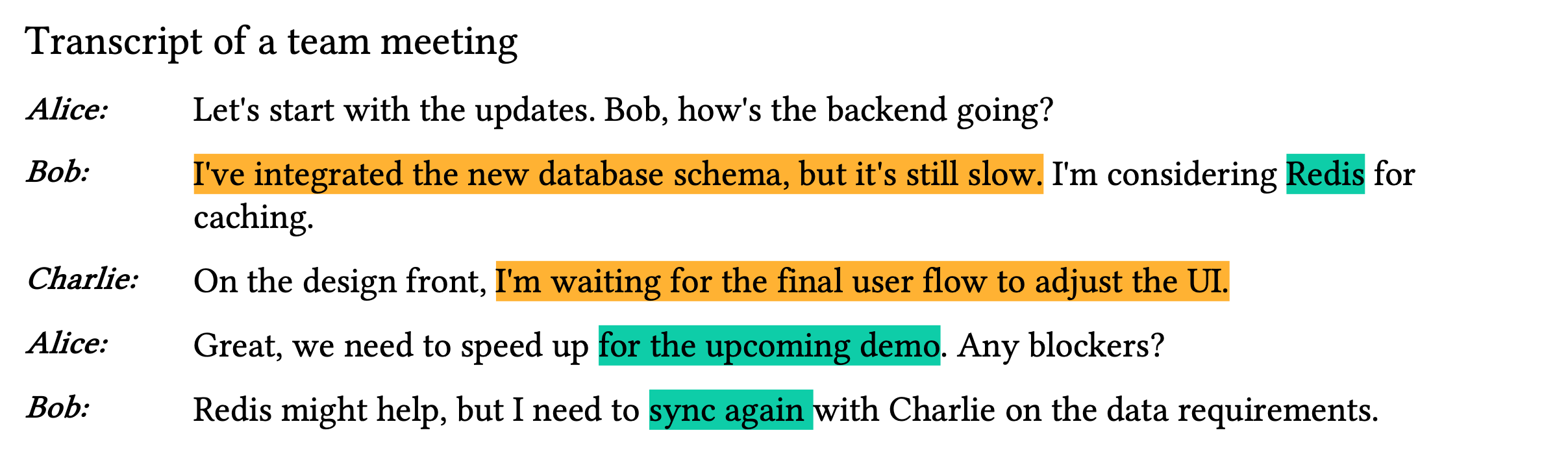}
    \caption{Example conversation to demonstrate the comprehension challenge including direct and implied content. Directly stated (\colorbox{myorange}{orange}) are the explicit updates by Bob and Charlie about their respective tasks. \colorbox{mygreen}{Green} marks implied content such as a deadline for a demo and a current focus on performance and finalizing features (unmentioned background knowledge), Redis implies an understanding of its role in performance enhancement (organizational knowledge and reference to prior discussions on data requirements).}
    \label{fig:challenge_understanding}
\end{figure}

\subsection{Comprehension}
\label{sec:challenge_understanding}

\noindent \textbf{Characteristics.}
The comprehension challenge involves accurately understanding and contextualizing utterances such that models can grasp and leverage these insights for summarization \shortcite{KhalifaBM21}.
This includes \textit{direct} and \textit{implied content}, e.g., unmentioned background knowledge, organizational knowledge, and prior discussions not explicitly referred to (examples in \Cref{fig:challenge_understanding}).
To infer and understand this content, the whole context of a dialogue is required, spanning both the \textit{local context}, focusing on neighboring words and sentences, and the \textit{global context}, covering the entire dialogue's informational flow \shortcite{ZhangHZJ19}.
Insufficient comprehension can lead to misunderstanding the content, omitting parts, or falsely presenting the information.

\noindent \textbf{Approaches.}
Transformer-based models struggle with comprehension and contextualization due to the underlying dependency parsing challenges, particularly with unclear sentence boundaries as they appear in spoken language, which can lead to catastrophic forgetting and result in unreliability in handling long-range dependencies \shortcite{LeeKHU21}.
Additionally, these models interpret the text literally, affecting their ability to grasp nuanced meanings \shortcite{RaiC20,WanLDS21}.
As these models often rely on syntactic relationships between words, they miss the nuanced understanding humans use to interpret implied meanings of sarcasm, irony, euphemisms, and similar pragmatic elements \shortcite{FazlyCS09}.
Consequently, this can lead to summaries that miss deeper meanings or implications, resulting in content that may be shallow or misleading \shortcite{zotero-18877}.

Despite its importance, our review reveals limited engagement from the dialogue summarization community to improve the contextualizing capability of models considering both the short and long context.
Current efforts primarily rely on Transformer models' self-attention capabilities to comprehend local and global contexts \shortcite{BeltagyPC20}.
Despite this reliance, the context-aware extract-generate framework \shortcite{zotero-18877} has been proposed to reduce the risk of missing long-range contexts in complex dialogues, which uses context prompts to capture the immediate details and the overall narrative of text spans.
It first pinpoints relevant text parts that guide the extraction of key information, which is then used as a prompt base for generating detailed summaries.

\subsection{Speaker}
\label{sec:challenge_speaker}

\begin{figure}[ht]
    \centering
    \includegraphics[width=0.9\linewidth]{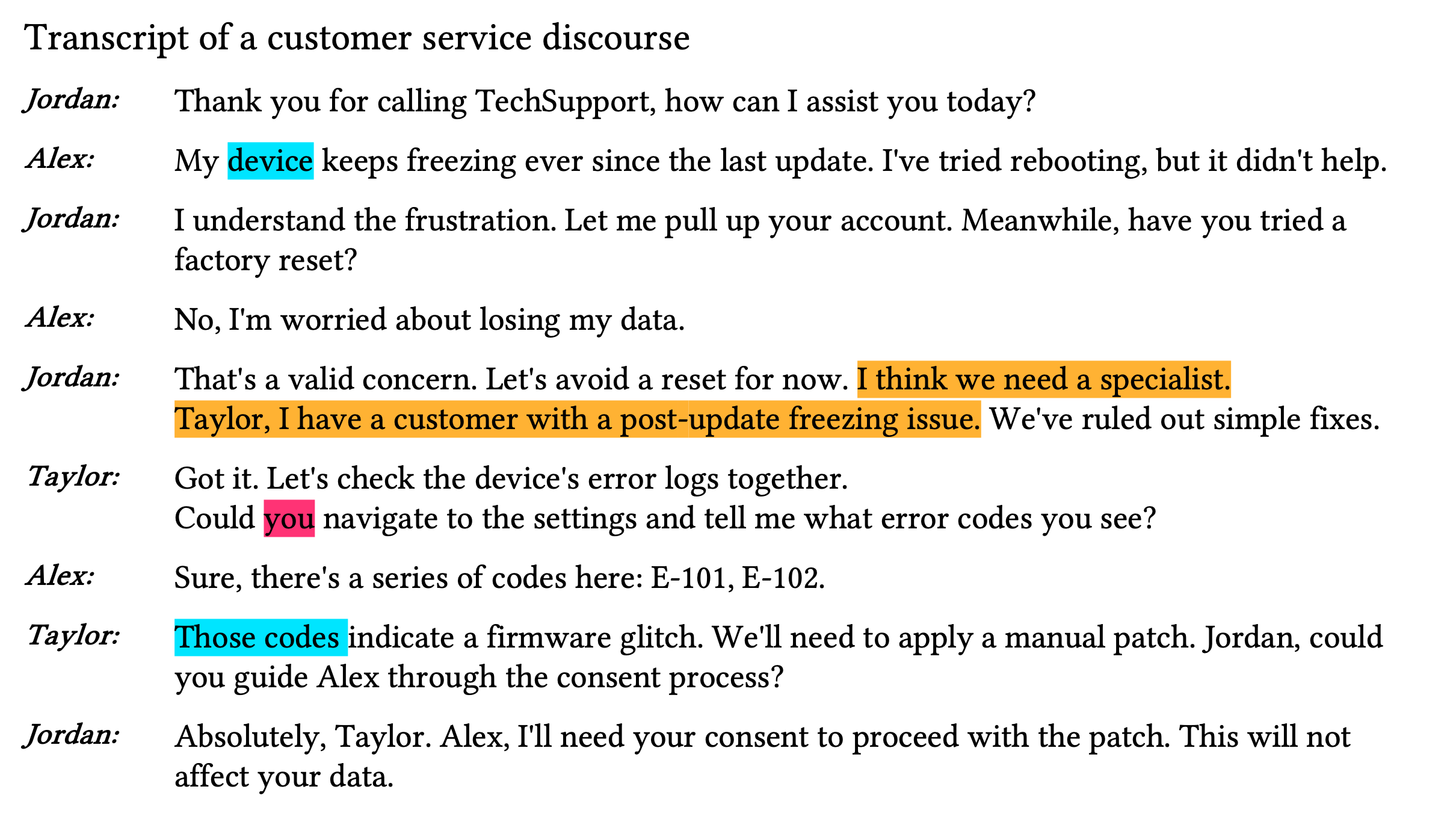}
    \caption{Example conversation of the speaker challenge with defined roles (Alex as customer, Jordan as customer support agent, Taylor as technical specialist). The dialogue shows a change of roles when Jordan calls Taylor (marked with \colorbox{myorange}{orange}), shifting from Jordan as a primary troubleshooter to a facilitator for Taylor's technical interventions. Further, multiple direct references to entities highlighted with \colorbox{myblue}{blue} (e.g., `device', `error codes') and indirect references to participants in \colorbox{myred}{red} (e.g., `you') are incorporated.}
    \label{fig:challenge_speaker}
\end{figure}

\noindent \textbf{Characteristics.}
Participants, their interactions, and how they influence the dynamics of a conversation are part of the speaker challenge.
This includes the individual \textit{participant roles} (e.g., `project manager,' `customer service agent', \shortciteauthor{LeiZYH21}, 2021b; \shortciteauthor{HuaDX22}, 2022), topic-dependent role changes (\shortciteauthor{KhalifaBM21}, 2021; \shortciteauthor{FengFQ22}, 2022a; \shortciteauthor{GuTL22}, 2022; \shortciteauthor{KumarK22}, 2022; \shortciteauthor{RennardSHV23}, 2022; \shortciteauthor{ShindeGSB22}, 2022; \shortciteauthor{AntonyASK23}, 2023), their \textit{interactions} (e.g., contributions depending on viewpoints, \shortciteauthor{BeckageHSM21}, 2021; \shortciteauthor{Li22}, 2022; \shortciteauthor{XieHZL22}, 2022; and on backgrounds, \shortciteauthor{LeeLWL21}, 2021; \shortciteauthor{ZouZHG21}, 2021; \shortciteauthor{GengZYQ22}, 2022), and \textit{references to entities and other participants} \shortcite{LeeYPS21,LiuC21,LiuC22,NarakiSH22}.
Examples of the speaker characteristics are shown in \Cref{fig:challenge_speaker}.
Failing to capture these speaker characteristics may lead to missing context, misaligned entity associations, and, ultimately, incorrect factual summaries. 
The significance of examining speaker dynamics and roles is well-recognized in other fields such as linguistics \shortcite{Meskill93} and is, together with the language challenge of the preceded analysis (\Cref{sec:challenge_language}), a typical hook to demonstrate and motivate techniques for dialogue summarization.

\noindent \textbf{Approaches.}
The speaker challenge stems from identifying participants, tracking their actions and entity references, and revealing participant relationships.
These aspects correspond to the difficulties of named entity recognition (NER), coreference resolution, and dependency parsing.
Despite advancements with Transformer-based models regarding NER, current techniques struggle with entity identification and categorization, particularly in unstructured texts and across diverse domains \shortcite{VajjalaB22,WangTZL22,Pakhale23}. 
This limitation is especially pronounced in realistic dialogue scenarios. 
Coreference resolution poses difficulties for techniques in accurately identifying mentions and contextual links \shortcite{QuanXWH19} and understanding the linguistic structures \shortcite{IoannidesJSN23}, extending the understanding challenge (\Cref{sec:challenge_understanding}).

Works typically use dense vectors \shortcite{AsiWEG22,NarakiSH22} to accurately represent speaker roles in dialogue and act as labels for utterances \shortcite{GanZKY21,QiLFL21}, offering insights into each speaker's function within the dialogue.
Speaker vectors are either randomly initialized and trainable \shortcite{ZhuXZH20a,QiLFL21} or produced by small neural networks \shortcite{GanZKY21,NarakiSH22} and represent different roles, e.g., `industrial designer' \shortcite{ZhuXZH20a}, `judge' \shortcite{DuanZYZ19}, which are appended to the embedding of the speaker's turn.
For capturing speaker dynamics and resolving coreferences, graph-based models are an established approach, e.g., representing each speaker's main ideas and their discourse, alongside their interactions, reflecting inner- and inter-speaker structures \shortcite{LeiZYH21,HuaDX22}.

Some other popular techniques involve enhancing Transformers' self-attention to analyze intra- and inter-speaker dynamics \shortcite{LeiYZH21,LiuSC21}, additional training tasks \shortcite{LeeYCJ21,GengZYQ22,Zhou23}, and pre-and post-processing techniques to replace pronouns and correct mistakes \shortcite{JoshiCLW20a,LeeLWL21,FangZCD22,LiuC22} for coreference resolution.

\begin{figure}[ht]
    \centering
    \includegraphics[width=0.9\linewidth]{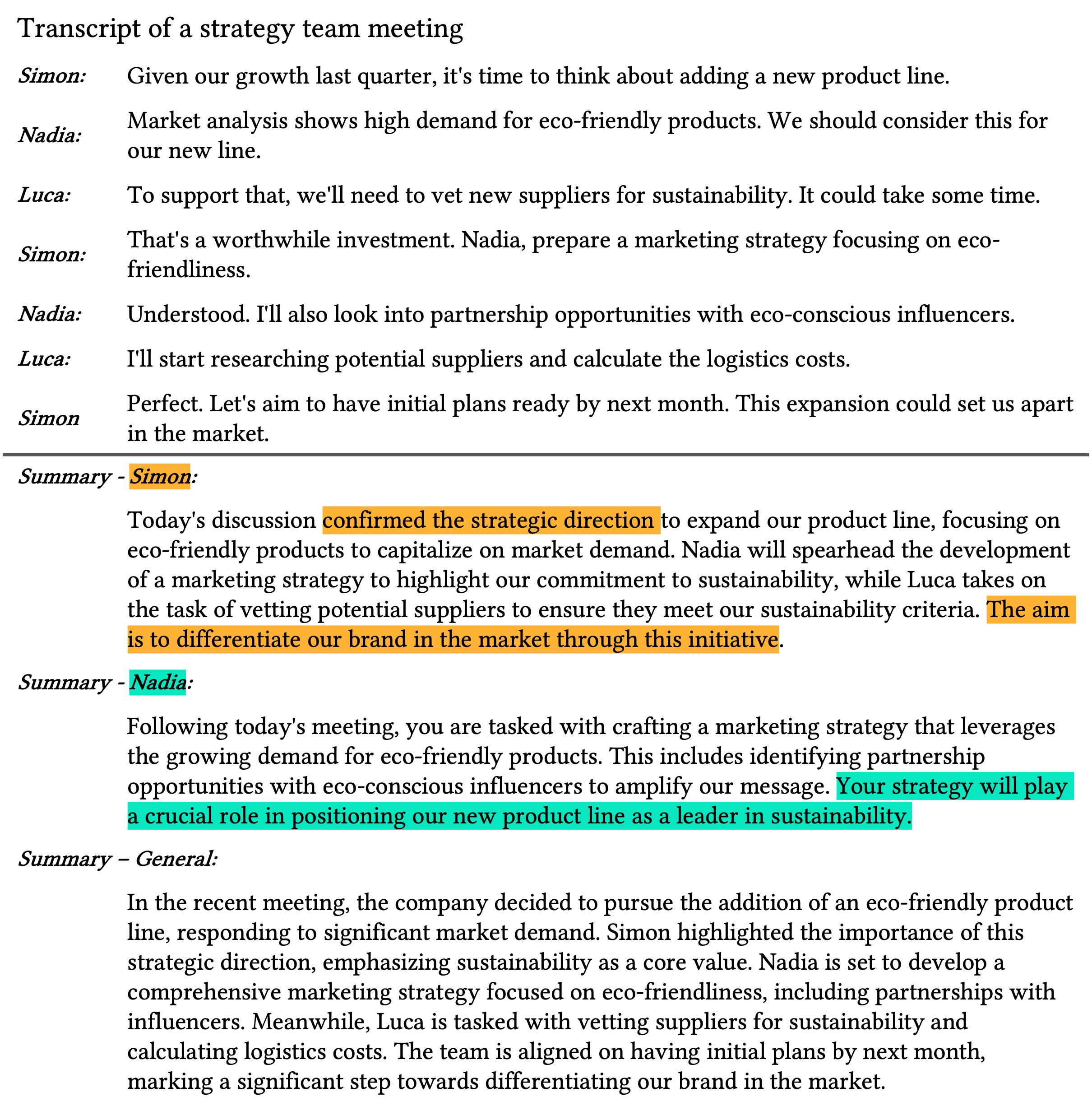}
    \caption{Example of the salience challenge with a conversation being summarized for a general audience and tailored summaries for participants with a specific focus on content relevant to them.}
    \label{fig:challenge_salience}
\end{figure}

\begin{figure}[ht]
    \centering
    \includegraphics[width=0.9\linewidth]{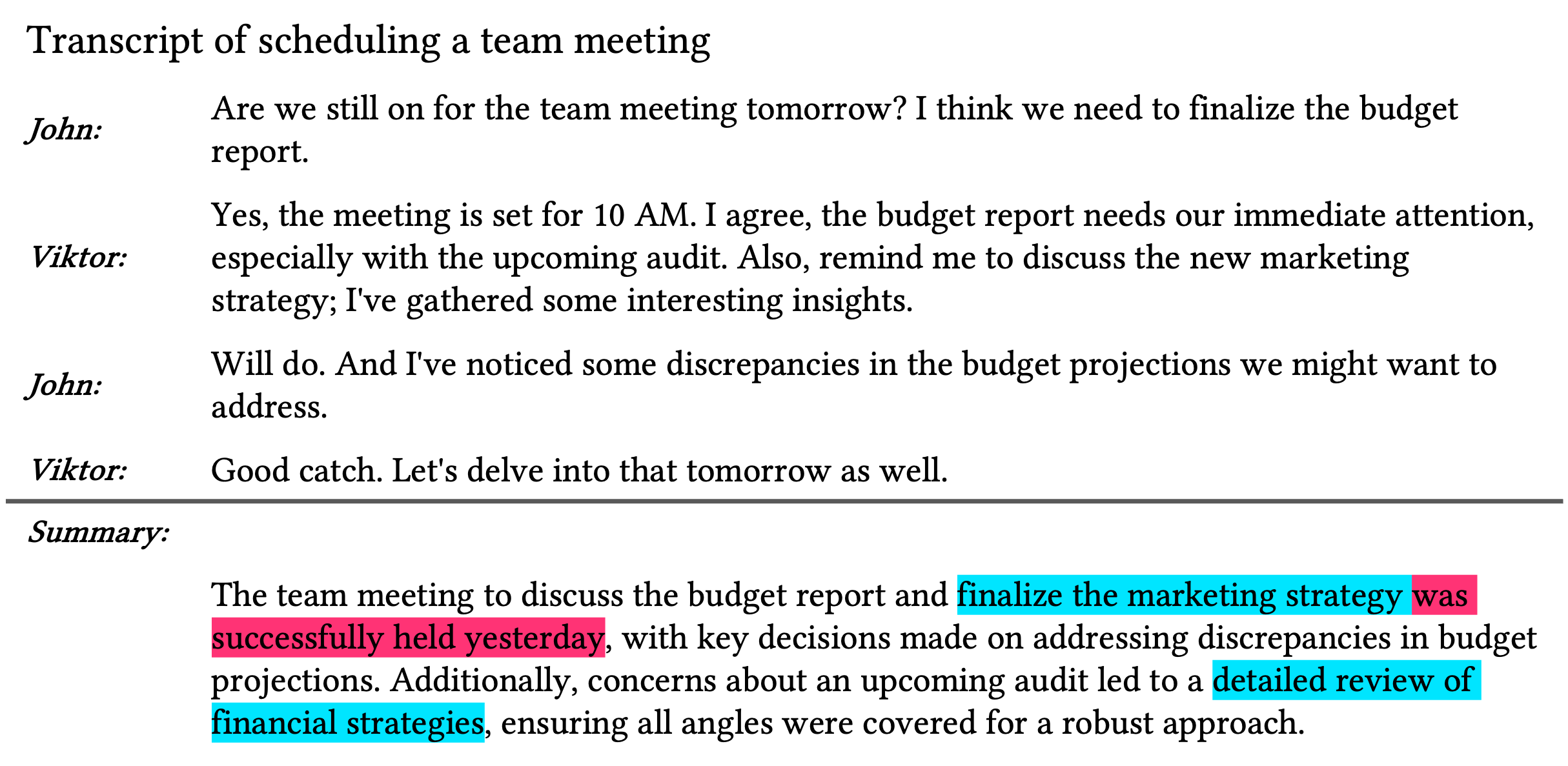}
    \caption{Example of the factuality challenge where the generated summary exhibits extrinsic errors through hallucinating events (\colorbox{myblue}{blue} background) such as the finalization of a marketing strategy and a detailed review of financial strategies. The intrinsic error reporting the wrong date of the meeting is marked with \colorbox{myred}{red}.}
    \label{fig:challenge_factuality}
\end{figure}

\subsection{Salience}
\label{sec:challenge_salience}

\noindent \textbf{Characteristics.}
The salience challenge involves identifying and summarizing the most relevant information within a dialogue \shortcite{ChenY20}.
This includes \textit{identifying} content critical for insightful summaries \shortcite{LiuZZC21}, where the understanding of critical shifted lately from content for general audience summaries to summaries aligning with the viewpoints, values, and ethics of specific participant \shortcite{LiuC21}, and \textit{locating} text spans containing relevant content \shortcite{LeiZYH21}.
\Cref{fig:challenge_salience} demonstrates the differences between personalized and generally applicable summaries, as the generated texts for Simon and Nadia are tailored to their role, stands, and tasks, whereas the general summary provides a neutral overview.
In contrast to traditional summarization, where salience may be related to the position of sentences (positional salience), dialogues are more dynamic, and salience emerges from how ideas evolve over the conversation.
Failure to effectively locate the scattered information or understand what is considered salient can impact the depth and relevance of the generated summaries, focusing on unimportant content while omitting relevant information bits \shortcite{ChenY20,ZhangNYZ21a,ZouSTF23,RennardSHV23}.

\noindent \textbf{Approaches.}
The first hurdle for current models to the salience challenge is the request for subjective salience and ranking content's importance to align with the role and knowledge of the summary addressee.
This is difficult as there is no training task for inferring personal attributes and ranking information according to them.
Further, locating this content is difficult as salient information is typically scattered across multiple turns within the idiosyncrasies of spoken language \shortcite{LeiZYH21,FengFQ22,TanWSZ23}, which forms an extension of the language challenge (\Cref{sec:challenge_language}).

To nest the general understanding of what is considered salient into models, established strategies include additional training stages to better distinguish between salient and non-salient content.
\shortciteA{HuangSMX23} introduce the uncovered loss to point out if salient information is missing and the contrastive loss to distinguish between relevant and irrelevant sentences, respectively.
Negative samples, categorized as redundant (i.e., text with unnecessary utterances) and null (i.e., text with relevant utterances removed), help models prioritize important content \shortcite{LiuZXL22a,LiuZXL22}.
To better determine salient content for specific participants, works propose a fine-tuning task in which the topic or user perspective to be focused on is passed as an input along the dialogue text \shortcite{ChauhanRDG22}, apply question-driven pre-training \shortcite{PagnoniFKW23} and use dynamic prompts to direct model attention to key dialogue aspects \shortcite{ZhangLYF23}.
Additional strategies involve direct text manipulation \shortcite{JungSJC23}, personal named entity planning to concentrate on specific entities \shortcite{LiuC21}, adapting attention mechanisms \shortcite{LiXYW21}, and integrating human feedback \shortcite{ChenDY23} or graph structures  \shortcite{HuaDM23} to further tailor summaries to the dialogue's core information.

\begin{table*} [t]
\centering
\small
\begin{tabular}{p{2.5cm} p{11.5cm}}
\toprule
\textbf{Challenge} & \textbf{Related Errors} \\
\midrule
Language & incoherence, repetition, hallucination\\
Structure & incoherence, omission, lack of depth\\
Comprehension & lack of context understanding, omission, hallucination\\
Speaker & lack of context understanding, false coreference resolution, hallucination\\
Salience & irrelevance, lack of context understanding\\
Factuality & hallucination\\
\bottomrule
\end{tabular}
\caption{Definition of eight to-be-expected error types in dialogue summarization, based on existing meeting summarization related error types \shortcite{KirsteinWRG24}.}
\label{tab:expected_error_types}
\end{table*}

\subsection{Factuality}
\label{sec:challenge_factuality}
In contrast to the previous input-related challenges, the factuality challenge describes the problem of correcting false content in a generated summary caused by a non-robust architecture.
Factual incorrect content contains \textit{extrinsic errors}, such as hallucinating events not present in the original transcript, and \textit{intrinsic errors}, such as incorrect coreference resolution misrepresenting actual event details, and wrong conclusions stemming from negation \shortcite{WangZZC22a}.
Extrinsic and intrinsic errors are shown in a sample dialogue in \Cref{fig:challenge_factuality}.
Given the importance of ensuring summaries accurately reflect the conversation, research focuses on creating safety measures to maintain credibility.

\noindent \textbf{Approaches.}
Research interest predominantly targets post-processing approaches that rewrite the generated summary to correct factual errors.
Notable approaches are the usage of \glspl{llm} for error detection and correcting \shortcite{LiWGS23} and reformulating the summary to closely match the original dialogue's predictive capabilities for subsequent content \shortcite{FuZWL21}.
Other approaches explore the use of auxiliary tasks \shortcite{TangNWW22} to estimate the factual aspects and predict the missing aspects in summary \shortcite{GanZKY21}, losses such as encouraging a model to generate sentences about content not yet covered and differentiating factual from non-factual sentences \shortcite{LiuZXL22a,HuangSMX23}, and human feedback \shortcite{ChenDY23}.
Further, architecture-modifying approaches are explored, such as a new encoder to grasp dialogue states \shortcite{ZhaoZHZ21} or to handle graph structures, where dialogue events are captured and organized in a graph. 
A slot-driven beam search algorithm is used in a filling-in-the-blanks setup to give priority to generating salient elements in the summary \shortcite{ZhaoZXG21}, and a hierarchical approach builds on sub-summaries \shortcite{NairSK23}.
Another noteworthy technique is the generation of a sketch to structure the final summary along this plan \shortcite{WuLLS21}.
Negation remains a long-standing challenge for language models often neglected in computational studies \shortcite{HossainHPB20,ZhangZ21}.

\section{Datasets}
\label{Sec:datasets}
In this review, we collected data generation methods and downstream datasets from the selected publications.
We only consider datasets used at least five times in our analysis to avoid including resources with little community impact.
This setup yields 18 datasets that we group into six categories based on the dialogue subdomains they stem from: daily chat, online chat, meeting, screenplay, customer service, medical, and debates.
We also create an `others' category for datasets that do not fit neatly into these subdomain groups.

These datasets typically condense a short dialogue of about 1k tokens between natural persons into a third-person summary of about 100 tokens.
More complex scenarios, such as meetings, parliamentary debates, and TV series dialogues, involve around 10k to 20k tokens in the transcript and at least four participants.
Due to privacy concerns, a significant portion of currently publicly available datasets are reenactments based on actual events, e.g., by Amazon Mechanical Turk\footnote{https://www.mturk.com/} workers, or established datasets such as \textsc{SAMSum} \shortcite{GliwaMBW19a} are modified to generate new data, e.g., by changing the dialogue transcript to a third-person report \shortcite{BertschNG22}.
This underscores that the variety of existing datasets is comparably limited, and these datasets may not capture real-life scenarios.
The 18 available established datasets are detailed in \Cref{sec:dataset_daily_chat,sec:dataset_online_chat,sec:dataset_meeting,sec:dataset_customer_service,sec:dataset_debates,sec:dataset_medical,sec:dataset_screenplay,sec:dataset_other}.
Given the limited number of available datasets, we present methods for creating artificial datasets and strategies for optimizing the use of existing datasets in \Cref{sec:techniques_datasets_artificial}.

For this section, we consider 93 papers to identify the established datasets, excluding works that do not mention a dataset, datasets that are no longer publicly available, and datasets in a language other than English.
We did not assess dataset quality or perform a detailed analysis of the inherent challenges, leaving that for future research.
In \Cref{tab:datasets_summary}, we connect the datasets with their respective primary challenge, as noted by their creators, to provide an overview of the distribution of challenges in datasets.


\subsection{Daily Chat}
\label{sec:dataset_daily_chat}

\textit{\textsc{DialogSum}} \shortcite{ChenLCZ21a} integrates dialogues from \textsc{DailyDialog} \shortcite{LiSSL17b}, \textsc{Dream} \shortcite{SunYCY19a}, \textsc{Mutual} \shortcite{CuiWLZ20a}, and English-speaking practice websites, featuring two-speaker interactions across daily-life scenarios like work, leisure, and travel.
Annotators were given guidelines on writing summaries, such as length constraints (no longer than 20\% of conversation length). 
\textsc{DialogSum} includes 13k dialogues, with inputs of around 1k tokens and summaries of 130 tokens.

\subsection{Online Chat}
\label{sec:dataset_online_chat}

\textit{\textsc{SAMSum}} \shortcite{GliwaMBW19a} comprises 16k written online dialogues from messaging apps, which linguists craft asked to create conversations similar to those they write daily, reflecting the proportion of topics of their real-life messenger conversations.
The messages, each written by one person, contain chit-chats, gossip, arranging meetings, discussing politics, and consulting university assignments. They typically involve two speakers, with an average of 94 tokens per conversation, varying between three and 30 turns\footnote{A turn is a contribution made by a speaker in the form of a single utterance or a statement.}.
A subset of \textsc{SAMSum} has been adapted into formal third-person language \shortcite{BertschNG22} to help models transition from informal dialogue to edited text, addressing the linguistic gap between pre-training and downstream task (detailed in the language challenge, \Cref{sec:challenge_language}).

\textit{\textsc{Forum}} \shortcite{BhatiaBM14} consists of random samples of 100 threads from the online discussion forums on \textsc{Ubuntu} and \textsc{TripAdvisor}, with a total of 556 and 916 posts, respectively.
Two human evaluators created summaries of the discussion threads, resulting in two human-written summaries per sample. 

\textit{\textsc{Crd3}} \shortcite{RameshkumarB20} consists of dialogues from the `Critical Role Dungeon and Dragon' show with summaries collected from the Fandom wiki, featuring 159 episodes with an average dialogue length of 2550 turns.

\subsection{Meeting}
\label{sec:dataset_meeting}

\textit{\textsc{AMI}} \shortcite{CarlettaABF06b} contains business meetings on the product design process, detailing 137 staged meetings on designing a new remote control.
Participants are project managers, marketing experts, user interface, and industrial designers.
The dataset includes transcripts and human summaries, with dialogues averaging 6k tokens over 535 turns and four speakers.
A modified version, \textsc{AMI-ITS} \shortcite{ManuvinakurikeSCN21}, offers additional annotations for incremental temporal summaries, which provide summaries for 100-second time durations on a subset of \textsc{AMI}.

\textit{\textsc{ICSI}} \shortcite{JaninBEE03c} has 59 academic group meetings with computer scientists, linguists, and engineers at the International Computer Science Institute (\textsc{ICSI}) in Berkeley, along with their summaries written by hired annotators.
The meetings have an average of 819 turns and 13k tokens with six speakers and are research discussions among students.

\textit{\textsc{QMSum}} \shortcite{ZhongYYZ21} introduces query-based summarization across diverse meeting domains, compiling 1.8k query-summary pairs from 232 meetings, encompassing product design (\textsc{AMI}), academic discussions (\textsc{ICSI}), and committee deliberations. 
The dialogues feature up to 13k tokens and six speakers.
The original task is to summarize the meeting given a stated question.
\textsc{MACSum-Dial} \shortcite{ZhangLYF23} is a modification of \textsc{QMSum}, designed for controllable summarization, highlighting mixed attributes such as length, attractiveness, and topic specificity.


\textit{\textsc{Elitr}} \shortcite{NedoluzhkoSHG22} features 120 English technical project meetings in computer science, with each transcript averaging 7k words, 730 turns, and six speakers.
The duration of the meetings varies from ten minutes to more than two hours, with an average of one hour long.
Meetings shorter than half an hour are exceptions, whereas meetings longer than two hours are topic-oriented mini-workshops, also rather occasional.
The original task of the dataset differs from abstractive summarization as a model is required to produce not an abstractive summary but a set of meeting minutes in bullet points.

\textit{\textsc{MeetingBank}} \shortcite{HuGDD23} contains meetings of city councils from six major U.S. cities occurring over the past decade.
It contains 1366 meetings spanning 3.6k hours, with a council meeting lasting an average of 2.6 hours and the transcript containing 28k tokens.

\subsection{Screenplay}
\label{sec:dataset_screenplay}

\textit{\textsc{MediaSum}} \shortcite{ZhuLMZ21} contains 463.6k media interview transcripts with summaries from National Public Radio (NPR, \shortciteauthor{MajumderLNM20}, 2020) and CNN, spanning a range of topics, including politics, news, crime, and economy.
The summaries are based on \textsc{NPR}'s interview overviews and CNN's topic descriptions, with the latter segmented at commercial breaks to match topics to corresponding interview segments.
On average, each transcript in this dataset contains 1.5k words, with summaries of around 11 words, typically involving seven speakers.

\textit{\textsc{Summscreen}} \shortcite{ChenCWG22} consists of 27k instances of TV series transcripts paired with human-written recaps sourced from \textsc{TV MegaSite} and \textsc{ForeverDreaming}.
The recaps of \textsc{ForeverDreaming} are based on community contributions stemming from Wikipedia and TVmaze.
Transcripts, typically including 28 speakers, average 6.6k tokens, and summaries around 380 tokens.


\subsection{Customer Service}
\label{sec:dataset_customer_service}

\textit{\textsc{TweetSumm}} \shortcite{FeigenblatGSJ21} contains 1.1k dialogues derived from Twitter customer support exchanges, each with three extractive and three abstractive human-written summaries. 
Originating from the \textsc{Kaggle Customer Support On Twitter} dataset, these real-world interactions span various industries, such as airlines and retail, averaging ten turns per dialogue and 36 tokens per abstractive summary.

\textit{\textsc{TodSum}} \shortcite{ZhaoZHZ21} is a customer service dataset based on \textsc{MultiWOZ} \shortcite{BudzianowskiWTC18a}, from which they select the five domains: restaurant, hotel, attraction, taxi, and train.
The dataset spans 10k samples with an average dialogue length of 187 utterances and 45 words in the summary.

\subsection{Medical}
\label{sec:dataset_medical}

\textit{\textsc{MTS-Dialog}} \shortcite{AbachaYFL23} is a collection of 1.7k simulated doctor-patient conversations with corresponding clinical notes serving as summaries, sourced from the public \textsc{Mtsamples} collection \shortcite{MoramarcoJSF21}.
This dataset encompasses various medical fields, including general medicine, neurology, orthopedics, dermatology, and immunology, adhering to the SOAP (Subjective, Objective, Assessment, Plan) note format. 
Dialogues average 18 sentences and 242 words, with summaries typically around 81 words.

\subsection{Debates}
\label{sec:dataset_debates}

\textit{\textsc{ADSC}} \shortcite{MisraAFW15} features sequences of two-party dialogue chains derived from the \textsc{Internet Argument Corpus} \shortcite{WalkerAAT12}, focusing on significant social and political topics like gun control, gay marriage, the death penalty, and abortion.
It includes 45 dialogues, each accompanied by five unique human-generated summaries, with each summary approximately 150 words in length.

\subsection{Others}
\label{sec:dataset_other}

\textit{\textsc{ConvoSumm}} \shortcite{FabbriRRW21} consolidates dialogues from four sources: New York Times comments, StackExchange, W3C emails, and Reddit, totaling 2k dialogues with 500 from each domain.
Crowdsourced workers from Amazon Mechanical Turk wrote the abstractive summaries with at most 90 tokens.
Inputs average 1.1k tokens, with summaries about 70 tokens in length.

\textit{\textsc{Ferranti}} \shortcite{zotero-18720} is a dataset designed for factual error correction in dialogue summarization, featuring 4k manually annotated items.
The original task is to evaluate the factuality of summaries and how to correct these summaries with minimal effort.
Drawing on \textsc{SAMSum} and \textsc{DialogSum}, \textsc{Ferranti} includes summaries produced by models like \textsc{BART} \shortcite{LewisLGG20} and \textsc{Unilm} \shortcite{DongYWW19a}.
Annotators assess these summaries for accuracy and identify errors, providing a focused tool for improving summary factualness.

\textit{\textsc{DialSum}} \shortcite{FangPZS} is based on the \textsc{VisDial} dataset \shortcite{DasKGS17}, where two participants discuss images from the \textsc{MSCOCO} dataset \shortcite{LinMBB15}, which features  $\sim$120k images.
Each image has five captions from five different annotators.

\newcolumntype{L}[1]{>{\raggedright\arraybackslash}p{#1}}
\begin{table}[t]
    \centering
    \small
    \begin{tabular}{L{1.7cm} L{2.3cm} L{2.7cm} L{4.5cm} c}
        \toprule
        \textbf{Type} & \textbf{Dataset} & \textbf{Challenge} & \textbf{Descriptive Tags} & \textbf{Usage} \\
        \midrule
        Daily Chat & \textsc{DialogSum} & Speaker & daily life scenarios & 26 \\
        \midrule
        Online & \textsc{SAMSum} & Speaker, Salience & messaging apps & 68 \\
        Chat &  \textsc{FORUM} & Salience  & threads & 5 \\
        &  \textsc{CRD3} & Language  & live-streamed show & 6 \\
        \midrule
        Meeting & \textsc{AMI} & Salience, Comprehension & staged business meetings & 63 \\
        &  \textsc{ICSI} & Salience, Language  & academic group meetings & 21 \\
        &  \textsc{QMSum} & Salience, Language  & query-based summarization & 18 \\
        &  \textsc{Elitr} & Salience  & technical project meetings in computer science & 7 \\
        & \textsc{MeetingBank} & Structure & parliament meetings & 7 \\
        \midrule
        Screenplay & \textsc{MediaSum} & Structure & media interview & 15 \\
        &  \textsc{SummScreen} & Speaker, Salience  & TV series transcripts & 8 \\
        \midrule
        Customer & \textsc{TweetSumm} & Language & Twitter customer support & 7 \\
        Service &  \textsc{TODSum} & Language, Factuality  & open-domain task-oriented dialogues & 6 \\
        \midrule
        Medical & \textsc{MTS-Dialog} & Comprehension & simulated doctor-patient conversations with clinical notes & 5 \\
        \midrule
        Debates & \textsc{ADSC} & Comprehension & two-party dialogues on social and political topics & 5 \\
        \midrule
        Others & \textsc{ConvoSumm} & Salience & dialogues from comments, emails, and threads & 5 \\
        & \textsc{Ferranti} & Factuality & human assessed automatic summaries & 5 \\
        & \textsc{DialSum} & Comprehension & two-party discussion on images & 10 \\
        \bottomrule
    \end{tabular}
    \caption{Matching between established datasets and their primary reported challenge. \\ `Usage' indicates the number of papers reporting the dataset in their publication.}
    \label{tab:datasets_summary}
\end{table}

\subsection{Data Augmentation and Utility}
Besides highlighting datasets with a notable community interest, we summarize in this subsection the research on techniques to cope with the data scarcity in dialogue summarization, covering artificial generation techniques of datasets through text generation and curriculum learning strategies to use the available data more effectively.

\noindent \textbf{Artificial Dataset Generation.}
\label{sec:techniques_datasets_artificial}
As discussed earlier, dialogue summarization faces challenges due to the need for adaptability across various domains, structures, and speaker dynamics, which typically would be addressed by training models on diverse datasets \shortcite{FengFQ22}.
However, datasets' scarcity and small size, with utterances containing just two to ten turns in areas like customer service and medical, restrict model training for more general applications.

Creating real-world datasets is costly and may conflict with data security and information communication policies, so artificially generated datasets are explored \shortcite{AbachaYFL23}.
Methods range from simple heuristic-based weak labeling, such as selecting the leading or longest utterance as proxy summary \shortcite{SznajderGLJ22}, to paraphrasing with updating the summary to maintain coherence \shortcite{LiuMSN22,wahle-etal-2022-large,WahleGR23}, to random alterations (e.g., swapping, deletion, dialogue-act-guided insertions) and changing conversation structures \shortcite{ChenY21,ParkSL22}.
The most common method is to use language models to create ground truth summaries either directly \shortcite{AsiWEG22,NairSK23,ZhouLCL23,ZhuQL23} or after training them on human summarization patterns through few-shot learning \shortcite{ChintaguntaKAK21}.

\noindent \textbf{Techniques to Maximize Dataset Utility.}
\label{sec:techniques_datasets_efficient}
As the training of Transformer-based models requires extensive data to generalize across various topics and dialogue formats \shortcite{FuZWL21}, researchers have explored ways to use the limited data available more effectively.
A key strategy involves a prompt-based curriculum learning strategy that gradually increases the degree of prompt perturbation (e.g., word swapping, content cutting) to improve the generalization ability of models \shortcite{LiWLD22}.
Another variation are dynamic prompts, which select best-fitting few-shot samples considering dialogue content, size, and speaker number \shortcite{ProdanP22}.
Also explored are prompt transfer techniques from related dialogue domains \shortcite{XieYWW23} to bolster the usage of dialogue state information (i.e., data used to represent the underlying intentions and goals within a dialogue).
Prompts are further used to split inputs into domain-invariant and domain-specific content to enhance model generalization \shortcite{ZhaoZZH22,LiXCZ23}.
Further techniques include freezing most model parameters and training only specific parameters for domain adaption \shortcite{ChenLXY23,SuriMSS23,ZhuYWZ23}.

\section{Evaluation}
\label{Sec:evaluation}
Researchers have adopted metrics from related fields and introduced new ones designed to assess the effectiveness of dialogue summarization methods.
These metrics proxy for quality characteristics, such as coherence, fluency, factuality, and accuracy.
Our literature review identifies 15 metrics used more than twice in 93 papers.
We categorize these metrics into four groups: count-based (e.g., \textsc{ROUGE} \shortciteauthor{Lin04}, 2004), model-based (e.g., \textsc{BARTScore} \shortciteauthor{YuanNL21}, 2021), QA-based (e.g., \textsc{QAEval} \shortciteauthor{DeutschBR21}, 2021), and human evaluation.
The automatic metrics strive for alignment with human judgment.
However, the correlation between these two is weak for dialogue summarization, and automatic metrics sometimes reward low-quality texts \shortcite{GattK18}.
Due to these recognized limitations in fully capturing the nuances of summarization quality, human evaluation is considered the gold standard across most studies.
Recent analyses \shortcite{GaoW22,KirsteinWRG24} conclude that established automatic metrics may work for a superficial understanding of a model's performance but do not align well with human judgment in discerning error nuances.
These metrics show individual weaknesses where they individually do not adequately reflect occurrence across all error types (e.g., of hallucinated content) or cannot show the impact on the quality in their scores (e.g., when information is omitted).
Hence, a composite metric of count-based, model-based, and QA-based metrics may be required for a more reliable automatic interpretation, with the individual metrics focusing on specific error types.

\Cref{sec:assessed_characteristics} details the characteristics typically evaluated by metrics and \Cref{sec:count_based,sec:model_based,sec:qa_based,sec:human_eval} further summarizes the 15 established metrics in dialogue summarization.
\Cref{tab:evaluation_summary} provides an overview of the identified metrics and their usage throughout the papers considered for this literature review.

\subsection{Assessed Characteristics }
\label{sec:assessed_characteristics}

Metrics assess the quality of a generated text, often in relation to a reference text. 
Low-quality text may contain more redundancy, incoherence, grammatical errors, poor structure, or inappropriate language, while high-quality text would closely align with the reference. 
We identify ten key characteristics discussed in the literature (in \textit{italic}) and organize them into four overarching groups: accuracy, content, readability, and style.

\noindent \textbf{Accuracy} is the core of a high-quality summary \shortcite{NetoFK02}, making it crucial to maintain the truth of the original text and ensure \textit{factuality}.
A summary must accurately reflect the facts, events, and details from the source without any distortion.

\noindent \textbf{Content} quality is mainly driven by \textit{relevance}.
A summary must contain the most relevant information, directly addressing the addressee's informational needs \shortcite{WilliamsTS14}.
This effort to overcome the salience challenge ensures that summaries prioritize the most important points for the reader.
\textit{Coverage} complements relevance by ensuring that all key topics and arguments from the dialogue are included, offering a comprehensive understanding without significant omissions \shortcite{MullickBRK24}.
The goal is to track both the salience and structure challenges with this.
\textit{Informativeness} goes a step further by selecting crucial information that conveys the depth of the input transcript, enabling readers to grasp the main points without needing to refer back to the original text \shortcite{SeeLM17c}.

\noindent \textbf{Readability} is essential for making summaries accessible and easy to grasp the key points quickly.
\textit{Coherence} ensures that information is presented logically with smooth transitions, making the summary easy to follow \shortcite{MullickBRK24}.
The \textit{structure and organization} of a summary further enhances its readability and facilitates information processing, ensuring that it is well-organized and logical \shortcite{CarbonellG98}.
\textit{Conciseness} \shortcite{BiswasI22} and \textit{non-redundancy} \shortcite{YangLSZ20} are also essential as a summary should contain only the essence of the input transcript without unnecessary details.

\noindent \textbf{Style} pertains to the text's perceived quality and formal presentation, contributing to its professional polish.
\textit{Consistency} in perspective, tense, and stylistic choices throughout the summary contributes to the overall perception of the text's professionalism \shortcite{KingSSW23}.
\textit{Fluency} focuses on grammar, vocabulary, and sentence structure and aims for a natural, easy-to-read summary free from phrasing errors, enhancing the information's overall clarity and accessibility \shortcite{KryscinskiKMX19}.

\subsection{Count-Based}
\label{sec:count_based}

Count-based metrics, including N-gram-based measures like \textsc{BLEU} \shortcite{PapineniRWZ02}, and \textsc{ROUGE} \shortcite{Lin04}, and statistical measures such as \textsc{Perplexity} \shortcite{JelinekMBB77}, are often static and rule-based algorithms that have a long history for evaluating text summarization.
However, due to their simplicity, they have been criticized for their limitations in capturing overall meaning, fluency, coherence, or factuality \shortcite{SaiMK22}. 

\textsc{BLEU} \shortcite{PapineniRWZ02} measures the precision of generated summaries against reference texts by examining the overlap of word sequences (typically 1- to 3-word N-grams, \shortciteauthor{YangDYC20}, 2019), emphasizing similarity to the original phrasing but risking a linear relationship to noise \shortcite{VaibhavSSN19}.

\textsc{ROUGE} \shortcite{Lin04} focuses on lexical similarity \shortcite{NgA15} and aims to capture the extent to which key content from the source is included in the summary, prioritizing content coverage through recall of N-gram overlap, thus attempting to address \textsc{BLEU}'s limitations by emphasizing content inclusion over mere precision.
A frequently used variation is \textsc{ROUGE-L}, which measures the longest common subsequence between a candidate and a reference.

\textsc{METEOR} \shortcite{BanerjeeL05} advances \textsc{BLEU} and \textsc{ROUGE} by incorporating both precision and recall, along with synonym matching for semantic analysis, thus enabling the capturing of the semantic similarity between a candidate and a reference.
Despite its more balanced approach and sentence-level focus, \textsc{METEOR}, like its predecessors, is prone to noise interference \shortcite{VaibhavSSN19}.

\textsc{chrF++} \shortcite{Popovic17} further expands the previous metrics by considering precision, recall, and F-score-based N-gram overlap at both character and word levels, offering a more granular analysis \shortcite{Popovic15}.

\textsc{CIDEr} \shortcite{VedantamZP15a} is a consensus-based metric that compares the similarity of a generated sentence against a set of human-written reference sentences.
The score is an aggregation of cosine similarity scores between the TF-IDF weighted N-grams of the generated and reference sentences, inherently capturing precision, recall, grammaticality, and salience \shortcite{LiL21a,FabbriKMX21a,LuWWJ22}.

\textsc{Perplexity} \shortcite{JelinekMBB77} diverges from the previous methods by the statistical approach of gauging a model's uncertainty in predicting word sequences, with lower scores indicating better alignment with expected language patterns.
This measure is, in contrast to the previous, an intrinsic evaluation metric that directly evaluates the language modeling objectives through text predictability rather than assessing the performance on the downstream task.
Therefore, additional metrics are required for a holistic evaluation of text generation quality.

\subsection{Model-Based}
\label{sec:model_based}

With advancements in language models, there is an increasing focus on model-based evaluation metrics due to their higher correlation with human judgment, though count-based metrics remain popular.
Model-based metrics encompass a wide range of approaches, including those that calculate semantic similarity (e.g., \textsc{BERTScore}, \textsc{MoverScore}), text generation likelihood (e.g., \textsc{BARTScore}), and entailment probability (e.g., \textsc{FactCC}). 
These metrics typically represent text in a latent space using pre-trained embeddings or contextual representations, aiming to provide a more nuanced assessment by focusing on semantic similarity, likelihood of text generation, and factual consistency.
While these metrics have the potential to adapt to evolving language use (e.g., distribution drifts) \shortcite{SellamDP20}, they can still be error-prone \shortcite{JiLFY23a}, slower than count-based metrics, and may not directly measure specific characteristics (as outlined in \Cref{sec:assessed_characteristics}), making it challenging to discern which particular aspect influences their scores.

\textsc{BERTScore} \shortcite{ZhangKWW20} leverages \textsc{BERT} embeddings \shortcite{DevlinCLT19a} to assess textual similarity.
The metric first contextually embeds the reference and candidate texts, then constructs a similarity matrix through pairwise cosine similarities on the token level.
The final score is computed by greedily selecting the highest similarity scores and calculating the harmonic mean of precision and recall, enabling the metric to capture semantic nuances beyond surface-level matching \shortcite{ZhaoPLG19}.

\textsc{MoverScore} \shortcite{ZhaoPLG19} was introduced concurrently to \textsc{BERTScore} following a similar approach, as it also leverages BERT embeddings but instead considers the distance between reference and candidate text, hence employing the \textsc{Word Mover's Distanc} \shortcite{KusnerSKW15}, a special case of the \textsc{Earth Mover's Distance} \shortcite{RubnerTG00}, to measure semantic distance.
This change in measure allows \textsc{MoverScore} to map semantically related words from one sequence to one word in another sequence (many-to-one). 

\textsc{BARTScore} \shortcite{YuanNL21} evaluates the plausibility of generating a reference text from a given generated text, and vice versa, by calculating the log-likelihood of a sequence that a BART \shortcite{LewisLGG20} model would typically generate based on the given context.
This evaluation focuses on assessing both the fluency and the semantic accuracy. 

\textsc{BLEURT} \shortcite{SellamDP20} extends beyond mere embedding comparisons by incorporating a \textsc{BERT} model pre-trained on lexical- and semantic-level supervision signals and fine-tuned on human judgments, enabling it to make detailed assessments of text quality, including coherence and relevance.

\textsc{BLANC} \shortcite{VasilyevDB20a} leverages \textsc{BERT} to perform a fill-in-the-blank task, both with and without the generated summary, to assess how informative or helpful the generated text is.
The difference in prediction accuracy indicates the utility of the summary in helping understand the text.

\textsc{FactCC} \shortcite{KryscinskiMXS20a} evaluates a summary's factual consistency with its source document using an entailment classifier, scoring based on the proportion of sentences classified as factual consistent by a \textsc{BERT} model.
The metric struggles with complex inferences and subtle nuances beyond direct comparison.

\subsection{QA-Based}
\label{sec:qa_based}

The QA-based metrics we identify in the literature focus on the factual correctness of a summary by using external question-answering systems.
These metrics leverage pre-trained transformer-based models to generate questions and assess whether the summary contains the correct answers. 
Their effectiveness largely depends on the quality of the underlying QA systems, which may not always align perfectly with human evaluation standards.

\textsc{FEQA} \shortcite{DurmusHD20a} generates questions based on the summary content and verifies whether their answers can be found in the source document. 

\textsc{SummaQA} \shortcite{ScialomLPS19} derives questions from the source text and tries to answer these using the summary.
It expands the QA-based evaluation scope by incorporating various question types and emphasizing the summary's informativeness.

\textsc{QuestEval} \shortcite{DeutschBR21} combines the approaches of \textsc{FEQA} and \textsc{SummaQA}, thereby adopting a bidirectional strategy, generating questions from the summary to compare with the source text and vice versa.
This bidirectional evaluation offers a balanced and holistic assessment by considering included and omitted information in the summary.

\newcolumntype{L}[1]{>{\raggedright\arraybackslash}p{#1}}

\begin{table}[t]
    \centering
    \small
    \setlength{\extrarowheight}{3pt} 
    \begin{tabular}{L{1.7cm} L{1.7cm} L{2.5cm} L{6cm} c}
        \toprule
        \textbf{Type} & \textbf{Category} & \textbf{Metric} & \textbf{Descriptive Tags} & \textbf{Usage} \\
        \midrule
        \multirow{1}{*}{\shortstack[l]{Count \\ based}} & N-gram & \textsc{Bleu} & overlap, precision, multiple references & 21 \\
        &  & \textsc{Rouge}  & overlap, recall, one reference & 127 \\
        &  & \textsc{Meteor}  & overlap, precision and recall, one reference & 3 \\
        &  & \textsc{ChrF++}  & character-level, F1 score, one reference & 2 \\
        &  & \textsc{Cider}  & consensus-based, multiple references & 3 \\
        \cline{2-5}
        & Statistical & \textsc{Perplexity} & likelihood of word sequences & 3 \\
        \midrule
        \multirow{1}{*}{\shortstack[l]{Model \\ based}} & Hybrid & \textsc{Bertscore} & token-level, cosine similarity, one-to-one & 38 \\
        &  & \textsc{Moverscore} & token-level, mover distance, one-to-many & 3 \\
        &  & \textsc{Factcc} & entailment classifier, scores consistency & 9 \\
        \cline{2-5}
        & Trained & \textsc{Bleurt} & mimics human judgment & 7 \\
        &  & \textsc{Bartscore} & mimics human judgment, promptable & 13 \\
        &  & \textsc{Blanc} & fill-in-the-blank task with and without the summary & 2 \\
        \midrule
        \multirow{1}{*}{\shortstack[l]{QA \\ based}} & & \textsc{Feqa} & questions based on summary, answered through input & 38 \\
        &  & \textsc{Summaqa}  & questions based on input, answered through summary & 3 \\
        &  & \textsc{Questeval}  & combination of \textsc{Feqa} and \textsc{Summaqa} & 3 \\
        \midrule
        \multirow{1}{*}{\shortstack[l]{Human \\ Evaluation}} & Performance & \textsc{Likert Scale} & ordinal scale, e.g., 1 (worst) to 5 (best) & 6 \\
        &  & \textsc{Pairwise Comparison} & pick best sample out of two & 9 \\
        &  & \textsc{Best-Worst Scaling} & rank a list of samples & 5 \\
        \cline{2-5}
        & Agreement & \textsc{Krippendorff's Alpha} & disagreement, different data formats & 2 \\
        &  & \textsc{Cohen's Kappa} & agreements, categorical data, 2 raters & 3 \\
        &  & \textsc{Fleiss' Kappa} & agreement, nominal data, 2 raters & 2 \\
        \bottomrule
    \end{tabular}
    \caption{Relevant metrics and evaluation measures employed in dialogue summarization with more than twice reported use. `Usage' states the number of papers reporting the evaluation measure in their publication.}
    \label{tab:evaluation_summary}
\end{table}

\subsection{Human Evaluation}
\label{sec:human_eval}

Human evaluation is often considered the gold standard for assessing the quality of a summary.
It is typically performed through crowdsourcing annotators (e.g., Amazon Mechanical Turk) who label samples.
We identify established approaches for human evaluations on summary performance and annotator agreement.

\noindent \textbf{Performance.} 
Summary performance is typically evaluated using \textit{Likert} scales \shortcite{Likert32a,QaderJPL18,FengFQQ21,LuWWJ22}, which provide a simple rating system for quality assessment but lack detailed feedback on specific text issues \shortcite{DouFKS22a}.
Alternatives like \textit{pairwise comparison} \shortcite{ElderGOL18,LiuSLS21} and \textit{best-worst scaling} \shortcite{FinnL92,RotheNS20a} offer more nuanced evaluations, with best-worst scaling noted for its higher reliability \shortcite{KiritchenkoM17a}.
Despite these options, the Likert scale remains the predominant method.

\noindent \textbf{Agreement.}
The reliability of human-generated evaluations hinges on annotator agreement, also referred to as meta-evaluation \shortcite{YuanNL21}.
Despite its significance, the incorporation of agreement metrics is frequently overlooked.
We identify three established measures to determine inner-annotator agreements: \textsc{Krippendorff's Alpha}, \textsc{Cohen's Kappa}, and \textsc{Fleiss' Kappa}.
Their scores range from 0 (poor reliability) to 1 (perfect reliability), with reported scores typically between 0.65 and 0.85.
\textit{\textsc{Krippendorff's Alpha}} \shortcite{Krippendorff70} offers a flexible approach suitable for multiple raters, applicable to any level of measurement, i.e., nominal, ordinal, interval, or ratio, and accommodates both qualitative and quantitative assessments.
This measure is particularly valuable with a broad range of variables or when comparing agreements across different measurement scales but assumes that all annotators assess all samples.
\textit{\textsc{Cohen's Kappa}} \shortcite{Cohen60} is used to measure the agreement between two raters who each classify a set of items into mutually exclusive categories (e.g., yes/no).
\textit{\textsc{Fleiss' Kappa}} \shortcite{Fleiss71a} is also tailored for scenarios involving fixed-category classifications but extends \textsc{Cohen's Kappa} to multiple annotators, excelling in the evaluation of how consistently annotators categorize text segments or dialogue turns into predefined groups.

\section{Discussion}
\label{Sec:discussion}
Throughout this work, we provide an overview of the current state of challenges, datasets, and evaluation.
In Section \ref{sec:discussion_challenges}, we analyze the emerging trends such as \glspl{llm} on mitigating the individual challenges, finding that our challenge taxonomy remains up to date.
Section \ref{sec:discussion_datasets} discusses the increasing interest in datasets covering personalized summarization and realistic settings, while Section \ref{sec:discussion_evaluation} covers recent research on improving evaluation metrics.

\subsection{Remarks on Challenges}
\label{sec:discussion_challenges}
In this study, we have introduced the CADS taxonomy to organize the inherent challenges of abstractive dialogue summarization (i.e., language, speaker, salience, comprehension, structure, and factuality) and unify their underlying definitions.
While presented separately for clarity, the challenges are interdependent and influence each other.
Consequently, approaches should be researched to tackle the challenges simultaneously whenever possible.
Despite advancements in addressing the challenges, we observe that most of them are still a hurdle for models due to limitations in the Transformer components used in these models, the lack of capabilities in contextualization and few-shot adaption of the encoder-decoder backbone models, and missing mitigation techniques.

Since 2023, \gls{nlp} has seen a significant shift with the exploration of \glspl{llm} and optimized Transformer architecture components (e.g., ring attention, \shortciteauthor{LiuZA23}, 2023, and multi-token generation, \shortciteauthor{GloeckleIRL24}, 2024), with their application to dialogue summarization being explored later \shortcite{ZhouRP23a,LyuPLB24,MullickBRK24}.
Initial studies indicate that \glspl{llm} match or exceed the performance of task-specific models like DialogLED \shortcite{ZhongLXZ22}, even with their limitations (e.g., hallucinations, salience) \shortcite{LaskarFCB23}.
\shortciteA{KirsteinWRG24} show that encoder-decoder models may perform better against the speaker challenge, particularly regarding coreference resolution, and struggle with the structure challenge.
\glspl{llm} handle the comprehension challenge but seem to be sub-optimal regarding robustness to language and speakers.
These early observations state \glspl{llm} as a noteworthy alternative to more traditional techniques (e.g., dialogue-style pre-training, \shortciteauthor{ZhongLXZ22}, 2022, graph structures to represent speaker structures, \shortciteauthor{HuaDX22}, 2022, and special losses tailored to determine salience, \shortciteauthor{HuangSMX23}, 2023).

Following, we discuss how the hurdles of the currently employed encoder-decoder models (identified in \Cref{Sec:progress}) align with challenges in \gls{nlp} and discuss techniques introduced to mitigate these challenges, thereby deriving how they can aid dialogue summarization.

\noindent \textbf{Bridging the language gap.}
Adapting established summarization models from structured pre-training data to the less formal nature of dialogue remains challenging.
\Glspl{llm}, especially those trained on non-edited text such as models from the \textsc{GPT} and \textsc{GEMINI} series, are promising in this regard \shortcite{LyuDXD24}.
Their few-shot learning capabilities allow for rapid adaptation to new linguistic patterns more effectively.
At the same time, the pre-training on a large and diverse corpus improves robustness in handling informal language, style variations, and repetition \shortcite{WangCPX24}.

\noindent \textbf{Long distance handling and dependency parsing.}
Recent progress in \glspl{llm} such as GPT-4, \textsc{Claude-3}, and \textsc{Phi-3} shows a significant extension of the processable context length to over 100k tokens, exceeding the context size of earlier models like LED \shortcite{BeltagyPC20}, which handle up to 16k tokens.
This increase in context allows for handling long dependencies without chunking, which is crucial for understanding complex dialogue structures, speaker dynamics, and content understanding.
While traditional models could theoretically handle such lengthy contexts, practical limitations arise from the input-length-dependent quadratic computational costs associated with Transformer attention mechanisms \shortcite{VaswaniSPU17}.
To address these issues, techniques like sparse attention \shortcite{BeltagyPC20}, flash attention during the training stage \shortcite{DaoFER24} and ring attention for inference \shortcite{LiuZA23}, sharing weights across attention heads \shortcite{Shazeer19}, and conditional computation to reduce the overall computational load \shortcite{AinslieLdO23} are proposed to manage large inputs.

\noindent \textbf{Generalizability.}
Modifying pre-trained models to handle the salience and structure challenges independent of the conversation's domain is essential for practical applications (e.g., customer service).
Established generalization techniques involving prompting \shortcite{MaZGT22,WangTMR22}, few-shot learning \shortcite{DangMLG22,QinJ22}, and meta-learning \shortcite{VilaltaD02,HospedalesAMS22} are incipient in dialogue summarization.
Meanwhile, \glspl{llm} build on these techniques to leverage patterns observed during the training stage and boost their generalization capabilities \shortcite{WilsonPF23}.
However, \glspl{llm} may still struggle with bias mitigation and ensuring broad applicability across various contexts when encountered words are not present in the training corpus or in examples \shortcite{BakkerCST24,TalatNBC22,WolfWAL24a}.

\noindent \textbf{Coreference resolution.}
Large-scale pre-trained and fine-tuned language models currently set the benchmark within \gls{nlp} for coreference resolution \shortcite{LiuMLC23}.
Challenges remain with the ambiguities in reference and context \shortcite{KhuranaKKS23}, and the usage of \glspl{llm} for this task.

\noindent \textbf{Hallucination reduction.}
Avoiding hallucinated content in generated text is a major challenge in \gls{nlp} tasks using \glspl{llm}, especially in culturally sensitive contexts \shortcite{JiLFY23a,McIntoshLSW23}.
Mitigation techniques include confidence penalty regularization \shortcite{LuBH21,LiuYL23} to reduce overconfidence and enhance accuracy and refinement methods to review and post-process a summary \cite{KirsteinRG24c}.

These advancements throughout \gls{nlp} suggest upcoming shifts in the key challenges of dialogue summarization.
\begin{itemize}
    \item The \textit{language} challenge can lose impact if few-shot learning is used to prompt \glspl{llm}, making it easier to apply document summarization techniques to dialogues directly without additional finetuning.
    \item We expect progress for the \textit{structure} challenge using \glspl{llm}, but robustness issues may persist due to issues with generalization and processing long texts.
    \item The \textit{comprehension} challenge, especially in grasping implied meanings, remains largely unsolved and holds significant research potential.
    Research from related \gls{nlp} fields such as sentiment analysis \shortcite{SrivastavaVKS20} suggests that current models still struggle to understand implied content.
    At the same time, research on retrieval-augmented generation \shortcite{BalaguerBCF24a} could enhance contextualization, helping models better grasp the context of conversations.
    \item For the hurdles implied in the \textit{speaker} challenge, i.e., coreference resolution and dependency parsing, \glspl{llm} are explored but have not yet succeeded due to robustness weaknesses (e.g., errors due to uncommon names, overseeing details in long inputs), signaling that further research is required.
    \item The \textit{salience} challenge builds on personalizing summaries and understanding long discussions, which both are gaining attention lately.
    Due to the novelty, the personalized summaries require more research on techniques and benchmarking datasets.
    \item The ongoing issue of hallucination in language models continues to pose a challenge in ensuring the \textit{factuality} in generated summaries.
\end{itemize}

\subsection{Trends in Datasets}
\label{sec:discussion_datasets}
\noindent \textbf{Personalized Summarization.}
In contrast to the typically general summaries, personalized summaries tailored to the users' needs are becoming increasingly popular \shortcite{TepperHBR18,BhatnagarKK23,KirsteinRKG24}.
Recent advancements have explored cross-attention and decoder self-attention to enhance role-specific information capturing \shortcite{LinZXZ22}.
By integrating detailed personal attributes, models can better understand each participant's background and motivations, leading to more targeted summaries.
Dataset-wise, only a few, such as the Chinese CSDS \shortcite{LinMZX21}, incorporate personalized summaries, whereas established English datasets typically provide only a single, general summary.

\noindent \textbf{Realistic Datasets.} 
As models improve in few-shot learning and do not necessarily require finetuning, the reliance on large-scale datasets is shrinking.
Considering this trend, we expect a rise in smaller datasets, which allow for high-quality, realistic examples.
These examples may be sufficient to adapt models to new domains, make approaches more robust, and test them against realistic scenarios.

\subsection{Trends in Evaluation}
\label{sec:discussion_evaluation}
Evaluation metrics, vital for indicating performance and comparing new techniques, are currently mostly borrowed from related \gls{nlp} tasks such as translation or text generation (\Cref{Sec:evaluation}).
However, the effectiveness of the established metrics (e.g., ROUGE for dialogue summarization as identified in \Cref{tab:evaluation_summary}) is limited \shortcite{GaoW22,KirsteinWRG24}, and while it can serve as a proxy \shortcite{WangMAK22b} it provides insufficient insights into the true efficacy of new techniques.
The also popular model-based metrics (e.g., BARTScore) seem unable to align well with human judgments \shortcite{GaoW22} for dialogue summarization.
Recent developments in \gls{nlp} use \gls{llm}-based metrics such as GEMBA \shortcite{KocmiF23} and ICE \shortcite{JainKSF23} for evaluation \shortcite{NairSK23}, building on \glspl{llm}' advanced text comprehension.
This set of metrics thereby mimics human judgment and assesses common aspects (e.g., fluency, coverage, coherence) with continuous \shortcite{JainKSF23}, Likert scale \shortcite{Likert32a,ChiangL23a}, or probability scores \shortcite{FuNJL23}.
\gls{llm}-based metrics offer a promising direction for dialogue summarization evaluation due to their customizability \cite{KirsteinRG24b}, though this area remains largely unexplored.

\section{Final Considerations}
\label{Sec:final_considerations}
In this article, we reviewed works on Transformer-based abstractive dialogue summarization.
We unified the existing definitions and concepts of dialogue summarization into a taxonomy (CADS) of the field's six main challenge blocks: language, structure, comprehension, speaker, salience, and reliability-related factuality.
We demonstrated how these challenges appear in dialogue summarization and discussed why they occur for established encoder-decoder-based summarization systems.
We then grouped and highlighted techniques introduced since 2019 under the challenges they tackle.
Despite advancements, we observed that the field is still in its infancy and offers ample opportunities for further research.
We also listed datasets used in existing studies to illustrate the extent of data scarcity in dialogue summarization. Additionally, we evaluated common metrics and noted a strong reliance on the ROUGE metric, coupled with a lack of human evaluation reports, which raises doubts about the actual effectiveness of current techniques.
In our discussion, we evaluated how current \gls{nlp} techniques address these challenges and derived implications of using \glspl{llm} for the field.
We thereby identify a lack of exploring such new models and borrowing effective techniques from related fields to overcome the limitations currently holding back progress.
We conclude that while challenges like language may become less relevant, others, such as comprehension and factuality, still require more exploration.
We recommend that future work should further discuss and adapt this taxonomy as new challenges, techniques, datasets, and evaluation measures emerge and as current challenges may be solved naturally through technological innovation.
The considered literature can be found in the accompanying repository\footnote{https://github.com/FKIRSTE/LitRev-DialogueSum \label{repository_link}} which will be regularly updated.

In the following, we discuss the discuss limitations of evidence (Section \ref{sec:limitation_evidence}) and in our review process (Section \ref{sec:limitation_process}).

\subsection{Limitations of Evidence}
\label{sec:limitation_evidence}
Given our selection of papers on dialogue summarization, we found limitations in current works that may bias our findings.
First, the crawled research studies in dialogue summarization predominantly focus on English-language dialogues, which renders their performance in multilingual or cross-cultural contexts unknown.
While we frequently came across Chinese datasets, other languages are underrepresented.
As a consequence, we focus our work on English dialogue summarization.
Second, most datasets stem from specific sectors, such as customer service or business meetings, which may not reflect the various dialogue types in different environments.
Lastly, the lack of detailed human evaluation reporting across research works, including inter-annotator agreement scores, further complicates the assessment of new techniques' usefulness.

\subsection{Limitations of Review Process}
\label{sec:limitation_process}
In our systematic review of dialogue summarization, we encountered methodological challenges that we had to mitigate.
First, we focused our search on studies published in English and mainly using English datasets.
While this is a limiting factor, most research on dialogue summarization considers only English dialogues, making this a feasible approximation.
However, we encourage others to assess the issues we point out for other languages.
Second, despite considering multiple databases, the timing and methods of our search may have missed studies, thus not fully capturing the dialogue summarization field.
To mitigate this limitation, we updated our crawl several times throughout writing this article and retrieved works from two databases.
We considered both peer-reviewed and non-peer-reviewed papers, which risks including less credible studies or misinformation.
To mitigate the potential weaknesses of non-peer-reviewed works, we use the ranking from semantic scholar and DBLP, which should account for the paper's quality and evaluate the studies' quality based on adherence to established reporting standards of top-tier conferences.

\acks{This work was supported by the Lower Saxony Ministry of Science and Culture and the VW Foundation.
The Mercedes-Benz AG Research and Development supported Frederic Kirstein.
}


\vskip 0.2in
\bibliography{LitRev_JAIR, addition_copyedit}
\bibliographystyle{theapa}

\end{document}